\documentclass[12pt]{article}
\usepackage[letterpaper,margin=2cm]{geometry}

\usepackage{hyperref}       
\usepackage{url}            
\usepackage{booktabs}       
\usepackage{amsfonts}       
\usepackage{nicefrac}       
\usepackage{microtype}      
\usepackage{xcolor}         
\usepackage{graphicx}
\usepackage{amssymb}
\usepackage{amsmath}
\usepackage{float}
\usepackage{subcaption}
\usepackage{tablefootnote}
\usepackage{algorithm} 
\usepackage{algorithmicx}
\usepackage{algpseudocode} 
\usepackage{adjustbox}
\usepackage{subcaption}
\usepackage{todonotes}
\usepackage{natbib}[square]

\newcommand{\Tau}{\mathcal{T}}

\DeclareMathOperator*{\argmin}{arg\,min}
\DeclareMathOperator*{\expect}{\mathbb{E}}

\title{Understanding Transfer Learning and Gradient-Based Meta-Learning Techniques\thanks{Accepted at Machine Learning Journal, Special Issue on Discovery Science 2021}}

\author{Mike Huisman         \and
        Aske Plaat           \and
        Jan N. van Rijn
}

\date{Leiden Institute of Advanced Computer Science, Leiden University}

\begin{document}

\maketitle

\begin{abstract}
Deep neural networks can yield good performance on various tasks but often require large amounts of data to train them.
Meta-learning received considerable attention as one approach to improve the generalization of these networks from a limited amount of data.
Whilst meta-learning techniques have been observed to be successful at this in various scenarios, recent results suggest that when evaluated on tasks from a different data distribution than the one used for training, a baseline that simply finetunes a pre-trained network may be more effective than more complicated meta-learning techniques such as MAML, which is one of the most popular meta-learning techniques.
This is surprising as the learning behaviour of MAML mimics that of finetuning: both rely on re-using learned features. 
We investigate the observed performance differences between finetuning, MAML, and another meta-learning technique called Reptile, and show that MAML and Reptile specialize for fast adaptation in low-data regimes of similar data distribution as the one used for training. 
Our findings show that both the output layer and the noisy training conditions induced by data scarcity play important roles in facilitating this specialization for MAML. 
Lastly, we show that the pre-trained features as obtained by the finetuning baseline are more diverse and discriminative than those learned by MAML and Reptile.
Due to this lack of diversity and distribution specialization, MAML and Reptile may fail to generalize to out-of-distribution tasks whereas finetuning can fall back on the diversity of the learned features. 
\end{abstract}

\section{Introduction}\label{sec:1}

Deep learning techniques have enabled breakthroughs in various areas such as game-playing \citep{silver2016mastering,mnih2015human}, image recognition \citep{krizhevsky2012imagenet,he2015delving}, and machine translation \cite{wu2016google}. 
However, deep neural networks are notoriously \textit{data-hungry} \citep{lecun2015deep}, limiting their successes to domains where sufficient data and computing resources are available \citep{hospedales2020meta,huisman2021}. 
\textit{Meta-learning} \citep{schaul2010metalearning,schmidhuber1987evolutionary,thrun1998lifelong,brazdil2022metalearning} is one approach to reduce these limitations by learning efficient deep learning algorithms across different tasks.
By presenting the learning algorithm with different tasks, that presumably share similarities with the task of interest, the learning algorithm is presumed to be able to learn more efficiently than when it has to learn the task of interest from scratch. 
This approach involves two different time scales of learning: at the \emph{inner-level}, a given task is learned, and at the \emph{outer-level} the learning algorithm is improved over tasks by adjusting the hyperparameters.
Seminal approaches for this are MAML and Reptile.

While the field attracted much attention, recent results \citep{chen2019closer,tian2020rethinking,mangla2020charting} suggest that simply pre-training a network on a large dataset and \textit{finetuning} only the final layer of the network (the final layer) may be more effective at learning new image classification tasks quickly than more complicated meta-learning techniques such as MAML \citep{finn2017model} and Reptile \citep{nichol2018reptile} when the data distribution is different from the one used for training.
In contrast, MAML and Reptile often outperform finetuning when the data distribution is similar to the one used during training.
These phenomena are not well understood and surprising as \citet{raghu2020rapid} have shown that the adaptation behaviour of MAML resembles that of finetuning when learning new tasks: most of the changes take place in the final layer of the network while the body of the network is mostly kept frozen. 

In this work, we aim to find an explanation for the observed performance differences between MAML and finetuning. More specifically, we aim to answer the following two research questions:
\begin{enumerate}
    \item Why do MAML and Reptile outperform finetuning in \emph{within-distribution} settings?
    \item Why can finetuning outperform gradient-based meta-learning techniques such as MAML and Reptile \citep{nichol2018reptile} when the test data distribution diverges from the training data distribution?
\end{enumerate}
Both questions focus on the \textbf{few-shot image classification settings}.
We base our work on MAML, Reptile and finetuning, as these are influential techniques that have sparked a large body of follow-up methods that use the underlying ideas.
Since the questions that we aim to answer are inherently harder than just a simple performance comparison, answering them for the models that are at the basis of this body of literature will be the right starting point.
We think that developing a better understanding of these influential methods is of great value and can cascade further onto the more complex methods built on top of these. 

Based on our analysis of the learning objectives of the three techniques (finetuning, MAML, Reptile), we hypothesize that MAML and Reptile specialize for adaptation in low-data regimes of tasks from the training distribution, giving them an advantage in within-distribution settings.
However, since they may settle for initial features that are inferior compared with finetuning due to their negligence, or relative negligence, of the initial performance, they may perform comparatively worse when the test data distribution diverges from the training distribution.

The primary contributions of our work are the following. 
First, we show the importance of the output layer weights and data scarcity during training for Reptile and MAML to facilitate specialization for quick adaptation in low-data regimes of similar distributions, giving them an advantage compared with finetuning.
Second, we show that the pre-trained features of the finetuning technique are more diverse and discriminative than those learned by MAML and Reptile, which can be advantageous in out-of-distribution settings.\footnote{All code for reproducing our results can be found at \url{https://github.com/mikehuisman/transfer-meta-feature-representations}}

\section{Related work}

Meta-learning is a popular approach to enable deep neural networks to learn from a few data by learning an efficient learning algorithm. 
Many architectures and model types have been proposed, such as MAML~\citep{finn2017model}, the meta-learner LSTM~\citep{ravi2017optimization}, TURTLE~\citep{huisman2021stateless} and MetaOptNet~\citep{lee2019meta}. 
However, our understanding of newly proposed techniques remains limited in some cases. 
For example, different techniques use different backbones which raises the question of whether performance differences between techniques are due to new model-types or due to the difference in used backbones \citep{huisman2021}. 

\citet{chen2019closer} was one of the first that investigated this question by performing a fair comparison between popular meta-learning techniques, including MAML \citep{finn2017model}, on few-shot image classification benchmarks such as miniImageNet \citep{vinyals2016matching,ravi2017optimization} and CUB \citep{wah2011caltech}.
Their results show that MAML often outperforms finetuning when the test tasks come from a similar data distribution as the training distribution when using shallow backbones. 
When the backbone becomes deeper and/or the domain differences between training and test tasks increase, however, this performance gap is reduced and, in some cases, finetuning outperforms MAML.

In addition to these findings by \citet{chen2019closer}, \citet{tian2020rethinking} demonstrate that simply finetuning a pre-trained feature embedding module yields better performance than popular meta-learning techniques (including MAML) on few-shot benchmarks.
\citet{mangla2020charting} and \citet{yang2021free} further support this finding as they have proposed new few-shot learning techniques based on finetuning pre-trained networks which significantly outperform meta-learning techniques.

These performance differences between simple finetuning and more sophisticated techniques such as MAML may be surprising, as \citet{raghu2020rapid} found that the learning behaviour of MAML is similar to that of finetuning on image classification benchmarks.
More specifically, they compared the feature representations of MAML before and after task-specific adaptation, and show that MAML relies mostly on feature re-use instead of quick adaptation because the body of the network is barely adjusted, which resembles the learning dynamics of finetuning (see \autoref{sec:finetuning}).
\citet{collins2020does} compared the feature representations of MAML and the finetuning method (expected risk minimization) 
in linear regression settings and found that MAML finds an initialization closer to the hard tasks, characterized by their gentle loss landscapes with small gradients. 
We demonstrate a similar property: MAML has greater flexibility in picking an initialization as long as the post-adaptation performance is good.  

In this work, we aim to unite the findings of \citet{raghu2020rapid} and \citet{chen2019closer} by finding an answer to the question of why finetuning can outperform meta-learning techniques such as MAML and Reptile \citep{nichol2018reptile} in some image classification scenarios while it is outperformed in other scenarios (when using a shallow backbone or when train/test task distributions are similar).

\section{Background}
In this section, we briefly revise supervised learning and few-shot learning (the main problem setting used in this work) and describe finetuning, MAML, and Reptile in that context. 

\subsection{Supervised learning}

In the \emph{supervised learning} setting, we have a joint probability distribution over inputs $\mathbf{x}$ and corresponding outputs $\mathbf{y}$, i.e., $p(\mathbf{x}, \mathbf{y})$.
In the context of deep learning, the goal is to build deep neural networks that can predict for any given input $\mathbf{x}$ the correct output $\mathbf{y}$.
Throughout this paper, we assume that the neural network architecture $f$ is fixed and that we only wish to find a set of parameters $\mathbf{\theta}$ such that the network predictions $f_{\mathbf{\theta}}(\mathbf{x})$ are as good as possible. 
This can be done by updating the parameters $\mathbf{\theta}$ in order to minimize a loss function $\mathcal{L}_{\mathbf{x}_i, \mathbf{y}_i}(\mathbf{\theta})$ that captures how well the network parameterized by $\mathbf{\theta}$ is performing on input $\mathbf{x}_i$ and corresponding output $\mathbf{y}_i$. 
Here, network parameters $\theta$ are a weight matrix, where $\theta_{(i:j)}$ represent the weights of the $i^{th}$ until the $j^{th}$ layer (inclusive), where $0 < i < j \le L$.
Thus, under the joint distribution $p(\mathbf{x}, \mathbf{y})$, we wish to find 
\begin{align}
    \argmin_{\mathbf{\theta}} \expect_{\mathbf{x}_i, \mathbf{y}_i} \left[ \mathcal{L}_{\mathbf{x}_i, \mathbf{y}_i}(\mathbf{\theta}) \right], \label{eq:suplearning}
\end{align} where $(\mathbf{x}_i, \mathbf{y}_i)$ are sampled from the joint distribution $p(\mathbf{x}, \mathbf{y})$, i.e., $\mathbf{x}_i, \mathbf{y}_i \sim p(\mathbf{x}, \mathbf{y})$.

The most common way to approximate these parameters is by performing gradient descent on that loss function, which means that we update the parameters in the direction of the steepest descent
\begin{align}
    \mathbf{\theta}^{(t+1)} = \mathbf{\theta}^{(t)} - \alpha \nabla_{\mathbf{\theta}^{(t)}} \expect_{\mathbf{x}_i, \mathbf{y}_i} \left[ \mathcal{L}_{\mathbf{x}_i, \mathbf{y}_i}(\mathbf{\theta}^{(t)}) \right].\label{eq:graddes}
\end{align}
Here, $\nabla_{\mathbf{\theta}^{(t)}}$ is the gradient with respect to $\mathbf{\theta}^{(t)}$, $t$ indicates the time step, and $\alpha$ the learning rate or step size. 

\subsection{Few-shot learning}
\label{sec:fsl}

Few-shot learning is a special case of supervised learning, where the goal is to learn new tasks from only a limited number of examples, which is the main focus of this work and the techniques described below.
In order to enhance the learning process on a limited number of examples, the learner is presented with an additional set of tasks, so that it can learn about the learning process. 
Here, every task $\Tau_j$ consists of a data distribution $p_j(\mathbf{x}, \mathbf{y})$ and a loss function $\mathcal{L}$.
Since the loss function is often assumed to be fixed across all tasks, we henceforth use the term `task' to refer to the task data distribution. The loss function is often assumed to be fixed, and therefore, we henceforth mean data distribution $p_j(\mathbf{x}, \mathbf{y})$ or a sample from this distribution, depending on the context. 
One notable exception is made in \autoref{sec:toy}, where we abstract away from data distributions and define a task purely abstractly as a loss function.

Tasks are commonly sampled from a large meta-dataset $\mathcal{D} \backsim p_s(\mathbf{x}, \mathbf{y})$, which itself is a sample from a source distribution $p_s$. 
In the case of classification, this is often done as follows.
Suppose that the source distribution from which dataset $\mathcal{D}$ is sampled, is defined over a set of classes $\mathcal{Y} = \{ c_1,c_2,\ldots,c_n \}$.
Then, we can create tasks $\Tau_j$ by considering only a subspace of this source distribution corresponding to a subset of classes $S_j \subseteq \mathcal{Y}$.
The method can then be evaluated on tasks sampled from a disjoint subset of classes $S_m \subseteq \mathcal{Y}$, where $S_m \cap S_j = \empty$. 

Below, we give a concrete example of this procedure for the popular \textbf{$N$-way $k$-shot classification} setting \citep{finn2017model, vinyals2016matching, snell2017prototypical}.
Suppose that we have a classification dataset $\mathcal{D} = \{ (\mathbf{x}_1,\mathbf{y_1}), \allowbreak (\mathbf{x}_2,\mathbf{y_2}), \ldots, (\mathbf{x}_M,\mathbf{y_M})  \}$ of examples.
Then, we can create an $N$-way $k$-shot task $\Tau_j$ by sampling a subset of $N$ labels $S_j \subseteq \mathcal{Y}$, where $\vert S_j \vert=N$.
Moreover, we sample precisely $k$ examples for every class to form a training set, or \emph{support set} $D^{tr}_{\Tau_j}$, for that task, consisting of $ \vert D^{tr}_{\Tau_j} \vert = N \cdot k$ examples.
Lastly, the test set, or \emph{query set} $D^{te}_{\Tau_j}$, is obtained by sampling examples of the subset of classes $S_j$ from $\mathcal{D}$ that are not present in the support set. 
Techniques then train on the support set and evaluated on the query set in order to measure how well they have learned the task.
This is the problem setting that we will use throughout this work.

The deployment of an algorithm for few-shot learning is often done in three stages. 
In the \textit{meta-training} stage, the algorithm is presented with training tasks and uses them to adjust the prior, such as the initialization parameters.
After every X training tasks, the \textit{meta-validation} stage takes place, where the learner is validated on unseen meta-validation tasks.
Finally, after the training is completed, the learner with the best validation performance is evaluated in the \textit{meta-test} phase, where the learner is confronted with new tasks that have not been seen during training and validation. 
Importantly, the tasks between meta-training, meta-validation, and meta-test phases are disjoint. 
For example, in image classification, the classes in the meta-training tasks are not allowed to occur in meta-test tasks as we are interested in measuring the learning ability instead of memorization ability.
In regression settings, every task has its own ground-truth function (as in \autoref{sec:toy}). 
For example, every task could be a sine wave with a certain phase and amplitude \citep{finn2017model}.

\subsection{Finetuning}\label{sec:finetuning}

Achieving good generalization by minimizing the objective in \autoref{eq:suplearning} using gradient-based optimization often requires large amounts of data. 
This raises the question of how we can perform few-shot learning of tasks.
The transfer learning technique called \emph{finetuning} tackles this problem as follows. 
In the \emph{pre-training phase}, it minimizes \autoref{eq:suplearning} on a given source distribution $p_s(\mathbf{x}, \mathbf{y})$ using gradient descent as shown in \autoref{eq:graddes}. 
This leads to a sequence of updates that directly update the initialization parameters.
Then, it freezes the feature extraction module of the network: all parameters of the network through the penultimate layer, i.e., $\mathbf{\theta}_{(1:L-1)}$ where $L$ is the number of layers.
When presented with a target distribution $p_j(\mathbf{x}, \mathbf{y})$ from which we can sample fewer data, we can simply re-use the learned feature embedding module $f_{\mathbf{\theta}_{(1:L-1)}}$ (all hidden layers of the network excluding the output layer) for this new problem.
Then, in the \emph{finetuning phase}, it only trains the parameters in the final layer of the network $\mathbf{\theta}_{(L)}$ (the final layer).

By reducing the number of trainable parameters on the target problem, this technique effectively reduces the model complexity and prevents overfitting issues associated with the data scarcity in few-shot learning scenarios.
This comes at the cost of not being able to adjust the feature representations of inputs.
As a consequence, this approach fails when the pre-trained embedding module fails to produce informative representations of the target problem inputs.

\subsection{Reptile}

Instead of joint optimization on the source distribution, \emph{Reptile} \citep{nichol2018reptile} is a meta-learning algorithm and thus aims to learn how to learn.
For this, it splits the source distribution $p_s(\mathbf{x}, \mathbf{y})$ into a number of smaller task distributions $p_1(\mathbf{x}, \mathbf{y}), p_2(\mathbf{x}, \mathbf{y}),\ldots, p_n(\mathbf{x}, \mathbf{y})$, corresponding to tasks $\Tau_1, \Tau_2,\ldots \Tau_n$.
On a single task $\Tau_j$ for $j \in \{ 1,\ldots,n \}$, its objective is to minimize \autoref{eq:suplearning} under the task distribution $p_j(\mathbf{x}, \mathbf{y})$ using $T$ gradient descent update steps as shown in \autoref{eq:graddes}.
This results in a sequence of weight updates $\mathbf{\theta} \rightarrow \mathbf{\theta}^{(1)}_j \rightarrow \ldots \rightarrow \mathbf{\theta}^{(T)}_j$.
After task-specific adaptation, the initial parameters $\mathbf{\theta}$ are moved into the direction of $\mathbf{\theta}^{(T)}_j$
\begin{align}
    \mathbf{\theta} = \mathbf{\theta} + \epsilon \left( \mathbf{\theta}^{(T)}_j - \mathbf{\theta}  \right),\label{eq:defreptupdt}
\end{align}
where $\epsilon$ is the step size.
Intuitively, this update interpolates between the current initialization parameters $\mathbf{\theta}$ and the task-specific parameters $\mathbf{\theta}^{(T)}_j$.
The updated initialization $\mathbf{\theta}$ is then used as starting point when presented with new tasks, and the same process is repeated.  
It is easy to show that this update procedure corresponds to performing first-order optimization of the multi-step objective 
\begin{align}
    \argmin_{\mathbf{\theta}} \expect_{\Tau_j \sim p(\Tau)} \left(  \sum_{t=0}^{T-1} \expect_{\mathbf{x}_i, \mathbf{y}_i \sim p_j} \left[ \mathcal{L}_{t+1}(\mathbf{\mathbf{\theta}^{(t)}_j}) \right] \right),
\end{align}
where $\mathcal{L}_{t+1}$ is shorthand for the loss on a mini-batch sampled at time step $t$.

\subsection{MAML}

Another popular gradient-based meta-learning technique is MAML \citep{finn2017model}. 
Just like Reptile, MAML also splits the source distribution $p_s(\mathbf{x}, \mathbf{y})$ into a number of smaller task distributions $p_1(\mathbf{x}, \mathbf{y}), p_2(\mathbf{x}, \mathbf{y}),\ldots, p_n(\mathbf{x}, \mathbf{y})$, corresponding to tasks $\Tau_1, \Tau_2,\ldots \Tau_n$. 
On the training tasks, it aims to learn a weight initialization $\mathbf{\theta}$ from which new tasks can be learned more efficiently. 
However, instead of optimizing a multi-step loss function, MAML only optimizes the final performance after task-specific adaptation.
More specifically, this means that MAML is only interested in the performance of the final weights $\mathbf{\theta}^{(T)}_j$ on a task and not in intermediate performances of weights $\mathbf{\theta}^{(t)}_j$ for $t < T$.
In other words, MAML aims to find
\begin{align}
    \argmin_{\mathbf{\theta}} \expect_{\Tau_j \sim p(\Tau)} \left( \expect_{\mathbf{x}_i, \mathbf{y}_i \sim p_j} \left[ \mathcal{L}_{T}(\mathbf{\mathbf{\theta}^{(T)}_j}) \right]  \right).
\end{align}

To find these parameters, MAML updates its initialization parameters as follows
\begin{align}
    \mathbf{\theta} = \mathbf{\theta} - \beta \nabla_{\mathbf{\theta}} \mathcal{L}_{T+1}(\mathbf{\theta}^{(T)}_j), \label{eq:mamlupdate}
\end{align}
where $\beta$ is the learning rate and $\nabla_{\mathbf{\theta}} \mathcal{L}_{T+1}(\mathbf{\theta}^{(T)}_j) = \nabla_{\mathbf{\theta}_j^{(T)}} \mathcal{L}_{T+1}(\mathbf{\theta}^{(T)}_j) \nabla_{\mathbf{\theta}} \mathbf{\theta}^{(T)}_j$.
The factor $\nabla_{\mathbf{\theta}} \mathbf{\theta}^{(T)}_j$ contains second-order gradients and can be ignored by assuming that $\nabla_{\mathbf{\theta}} \mathbf{\theta}^{(T)}_j = I$ is the identity matrix, in a similar fashion to what Reptile does.
This assumption gives rise to \emph{first-order} MAML (fo-MAML) and significantly increases the training efficiency in terms of running time and memory usage, whilst achieving roughly the same performance as the \emph{second-order} MAML version \citep{finn2017model}.
In short, first-order MAML updates its initialization in the gradient update direction of the final task-specific parameters.
In this work, we focus on first-order MAML, as \citet{finn2017model} have shown this to perform similarly to second-order MAML.

\section{A common framework and interpretation}\label{sec:framework}

The three discussed techniques can be seen as part of a general gradient-based optimization framework, as shown in \autoref{alg:gengradopt}.
All algorithms try to find a good set of initial parameters as specified by their objective functions. 
The parameters are initialized randomly in line 1.
Then, these initial parameters are iteratively updated based on the learning objectives (the loop starting from line 2).

This iterative updating procedure continues as follows.
First, the data distribution is selected to sample data from (line 3). 
That is, finetuning uses the full joint distribution $p_s(\mathbf{x}, \mathbf{y})$ of the source problem, whereas Reptile and MAML select task distributions $p_j(\mathbf{x}, \mathbf{y})$ (obtained by sub-sampling a set of instances coming from a subset of labels from the full distribution $p_s$).
Next, we make $T$ task-specific updates on mini-batches sampled from the distribution $p$ that was selected in the previous stage (lines 4--8). 
Lastly, the initial parameters $\mathbf{\theta}$ are updated using the outcomes of the task-specific adaptation phase.

Note that in this general gradient-based optimization framework, all techniques update their initialization parameters based on a single distribution $p$ at a time.
One could also choose to use batches of distributions, or \emph{meta-batches}, in order to update the initialization $\mathbf{\theta}$.
This can be incorporated by using the average of the losses of the different distributions as an aggregated loss function. 

\begin{algorithm}[htb]
\caption{General gradient-based optimization: \colorbox{red!30}{finetuning} \colorbox{green!30}{reptile} \colorbox{blue!30}{MAML}} \label{alg:gengradopt} 
\begin{algorithmic}[1]
\State Randomly initialize $\mathbf{\theta}$
\While{not converged}
    \State Select data distribution $p=$ \colorbox{red!30}{$p_s$} \colorbox{green!30}{$p_j \sim p(\Tau)$} \colorbox{blue!30}{$p_j \sim p(\Tau)$}
    \State Set $\mathbf{\theta}^{(0)} = \mathbf{\theta}$
    \For{$t=0,...,T-1$}
        \State Sample a batch of data $\mathbf{x}, \mathbf{y} \sim p$
        \State Compute $\mathbf{\theta}^{(t+1)} = \mathbf{\theta}^{(t)} - \nabla_{\mathbf{\theta}^{(t)}}\mathcal{L}_{t+1}(\mathbf{\theta}^{(t)})$
    \EndFor
    \State Update $\mathbf{\theta}$ by \colorbox{red!30}{$\mathbf{\theta} = \mathbf{\theta}^{(T)}$} \colorbox{green!30}{\autoref{eq:defreptupdt}} \colorbox{blue!30}{\autoref{eq:mamlupdate}}
\EndWhile
\end{algorithmic}
\end{algorithm}

\autoref{tab:algorithms} gives an overview of the three algorithms. 
As we can see, finetuning only optimizes for the initial performance and does not take into account the performance after adaptation. 
This means that its goal is to correctly classify any input $\mathbf{x}$ from the source problem distribution $p_s$.
Reptile, on the other hand, optimizes both for initial performance, as well as performance after every update step. 
This means that Reptile may settle for an initialization with somewhat worse initial performance compared with finetuning, as long as the performance during task-specific adaptation makes up for this initial deficit.
MAML is the most extreme in the sense that it can settle for an initialization with poor initial performance, as long as the final performance is good. 

\begin{table}[tb]
    \centering
    \caption{Overview of the loss functions and corresponding focus of finetuning, Reptile, and MAML.}
    \label{tab:algorithms}
    \begin{tabular}{ccc}
        \toprule
         \textbf{Algorithm} & \textbf{Loss function} & \textbf{Focus} \\
         \midrule
         Finetuning & \( \displaystyle \expect_{\mathbf{x}_i, \mathbf{y}_i} \left[ \mathcal{L}_{\mathbf{x}_i, \mathbf{y}_i}(\mathbf{\theta}) \right] \) & Initial performance \\[0.3cm] 
         Reptile & \( \displaystyle \expect_{\Tau_j \sim p(\Tau)} \left(  \sum_{t=0}^{T-1} \expect_{\mathbf{x}_i, \mathbf{y}_i \sim p_j} \left[ \mathcal{L}_{t+1}(\mathbf{\mathbf{\theta}^{(t)}_j}) \right] \right) \) & Multi-step performance \\[0.6cm]
         MAML & \( \displaystyle \expect_{\Tau_j \sim p(\Tau)} \left( \expect_{\mathbf{x}_i, \mathbf{y}_i \sim p_j} \left[ \mathcal{L}_{T}(\mathbf{\mathbf{\theta}^{(T)}_j}) \right]  \right) \) & Final performance \\[0.3cm] 
         \bottomrule \\
    \end{tabular}
\end{table}

In short, Reptile and MAML can be interpreted as \emph{look-ahead algorithms} as they take the performance after task-specific adaptation into account whereas finetuning does not.
Moreover, fo-MAML relies purely on the look-ahead mechanism and neglects the initial performance while Reptile also takes the initial and intermediate performances into account. 
This means that MAML may outperform finetuning with a \emph{low-capacity} network (with the worst initial performance) where there is not enough capacity to store features that are directly useful for new tasks.
The reason for this is likely that finetuning will be unable to obtain good embeddings for all of the training tasks and does not have a mechanism to anticipate what features would be good to learn future tasks better.
MAML, on the other hand, does have this capability, and can thus settle for a set of features with worse initial performance that lends itself better for learning new tasks. 
In contrast, when we have \emph{high-capacity} networks with enough expressivity to store all relevant features for a task, finetuning may outperform MAML as it optimizes purely for initial performance without any additional adaptation, which can be prone to overfitting to the training data of the tasks due to the limited amount of available data. 
Lastly, one may expect Reptile to take place between MAML and finetuning: it works better than finetuning when using low-capacity backbones while it may be slightly worse than finetuning when using larger-capacity networks (but better than MAML). 

Although MAML focuses on the performance after learning, it has been shown that its learning behaviour is similar to that of finetuning: it mostly relies on feature re-use and not on fast learning \citep{raghu2020rapid}. 
This means that when a \emph{distribution shift} occurs, which means that the test tasks become more distant from the tasks that were used for training, MAML may be ill-positioned due to poor initial performance compared with finetuning which can fall back on more directly useful initial features.

\section{Experiments}
\label{sec:exps}

In this section, we perform various experiments to compare the learning behaviours of finetuning, MAML, and Reptile, in order to be able to study their within-distribution and out-of-distribution qualities that can help us answer the two research questions posed in \autoref{sec:1}. 
All experiments are conducted using single PNY GeForce RTX 2080TI GPUs. 
In order to study the question of why MAML and Reptile can outperform finetuning in within-distribution settings with a shallow Conv-4 backbone, we perform the following three first experiments.
Moreover, to investigate why finetuning can outperform MAML and Reptile in out-of-distribution settings, addressing our second research question, we perform experiment four listed below. 
\begin{enumerate}
    \item \textbf{Toy problem} (\autoref{sec:toy}) We study the behaviour of the algorithms on a \emph{within-distribution} toy problems where there are only two tasks without noise in the loss signals caused by a shortage of training data. This allows us to investigate the initializations that the methods settle for after training. This allows us to see why MAML and Reptile may have an advantage over finetuning in within-distribution settings.
    \item \textbf{The effect of the output layer} (\autoref{sec:output}) Finetuning removes the learned output layer and replaces it with a randomly initialized one when presented with a new task. MAML and Reptile, on the other hand, do not do this, and can directly start from the learned initialization weights for both the body and output layer of the network. To investigate whether this gives these two methods an advantage over finetuning in \emph{within-distrbution} few-shot image classification, we investigate the effect of replacing the learned output layers with randomly initialized ones before learning a new task. This allows us to determine the importance of having a learned weight initialization for the output layer and whether this is something that can explain the advantage of MAML and Reptile over finetuning in these settings. 
    \item \textbf{Specialization for robustness against overfitting} (\autoref{sec:noise}) Another difference between the methods is that finetuning is trained on regular mini-batches of data, whilst MAML and Reptile are trained explicitly for post-adaptation performance on noisy loss signals induced by the limited amount of available training data. To investigate the importance of explicitly training under noisy conditions, we study the performances of MAML and Reptile as a function of the number of examples present in the training condition. Here, the risk of overfitting is inversely related to the number of training examples $k$ per task.  
    \item \textbf{Information content in the learned initializations} (\autoref{sec:infcont}) Lastly, we investigate the within-distribution and out-of-distribution learning performances of finetuning, MAML, and Reptile, with three different backbones of different expressive power (Conv-4, Resnet-10, Resnet-18). More specifically, we propose a measure of broadness or discriminative power of the features and investigate whether this is related to the few-shot learning abilities of these methods to see whether the discriminative power of the three methods differ and can account for the potential superiority of finetuning in the out-of-distribution setting.   
\end{enumerate}

\begin{figure}[thb]
    \centering
    \begin{subfigure}[b]{0.49\linewidth}
    \includegraphics[width=\linewidth]{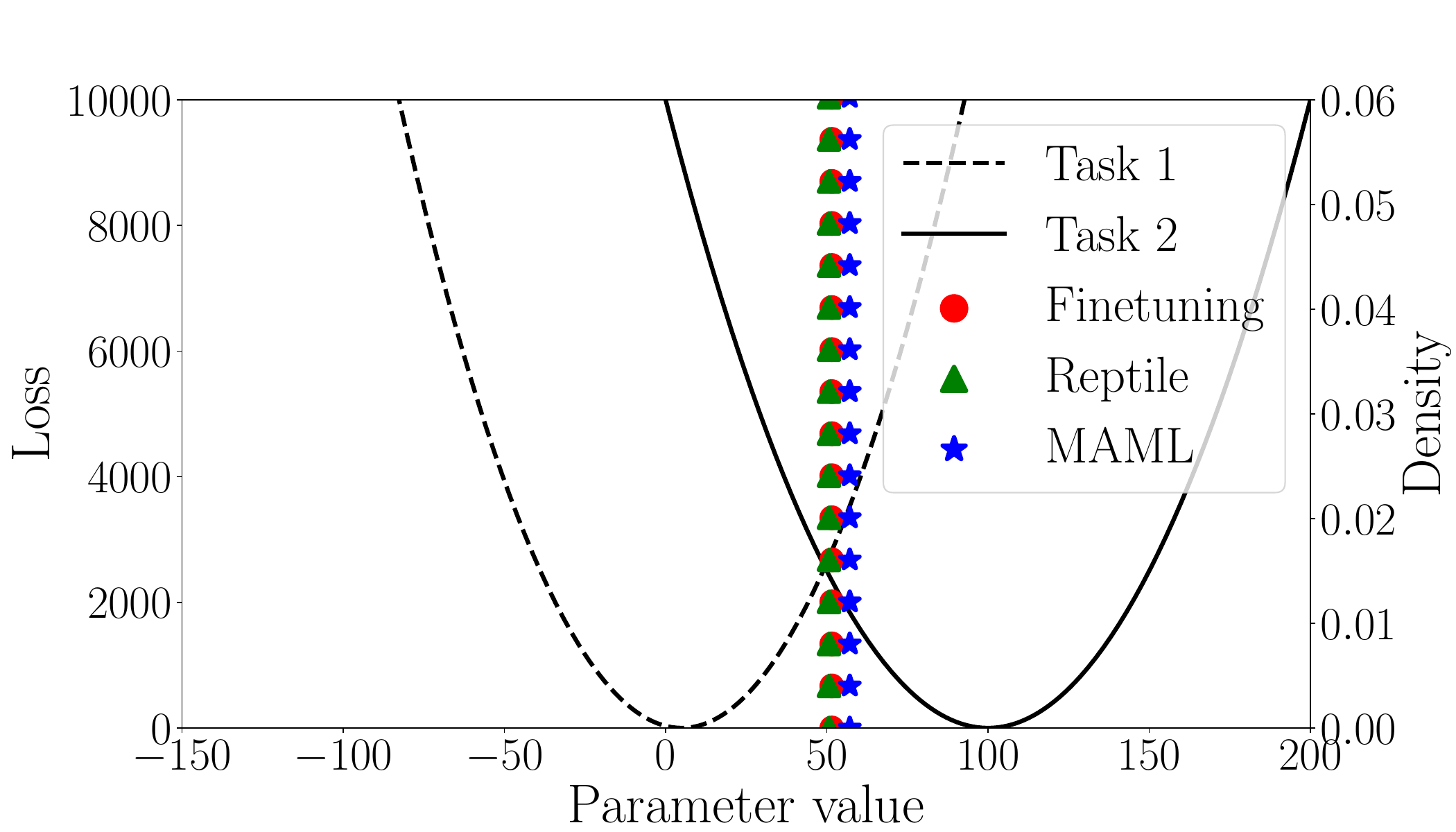}
    \caption{Scenario $a$, with $T=5$}
    \end{subfigure}
    \begin{subfigure}[b]{0.49\linewidth}
    \includegraphics[width=\linewidth]{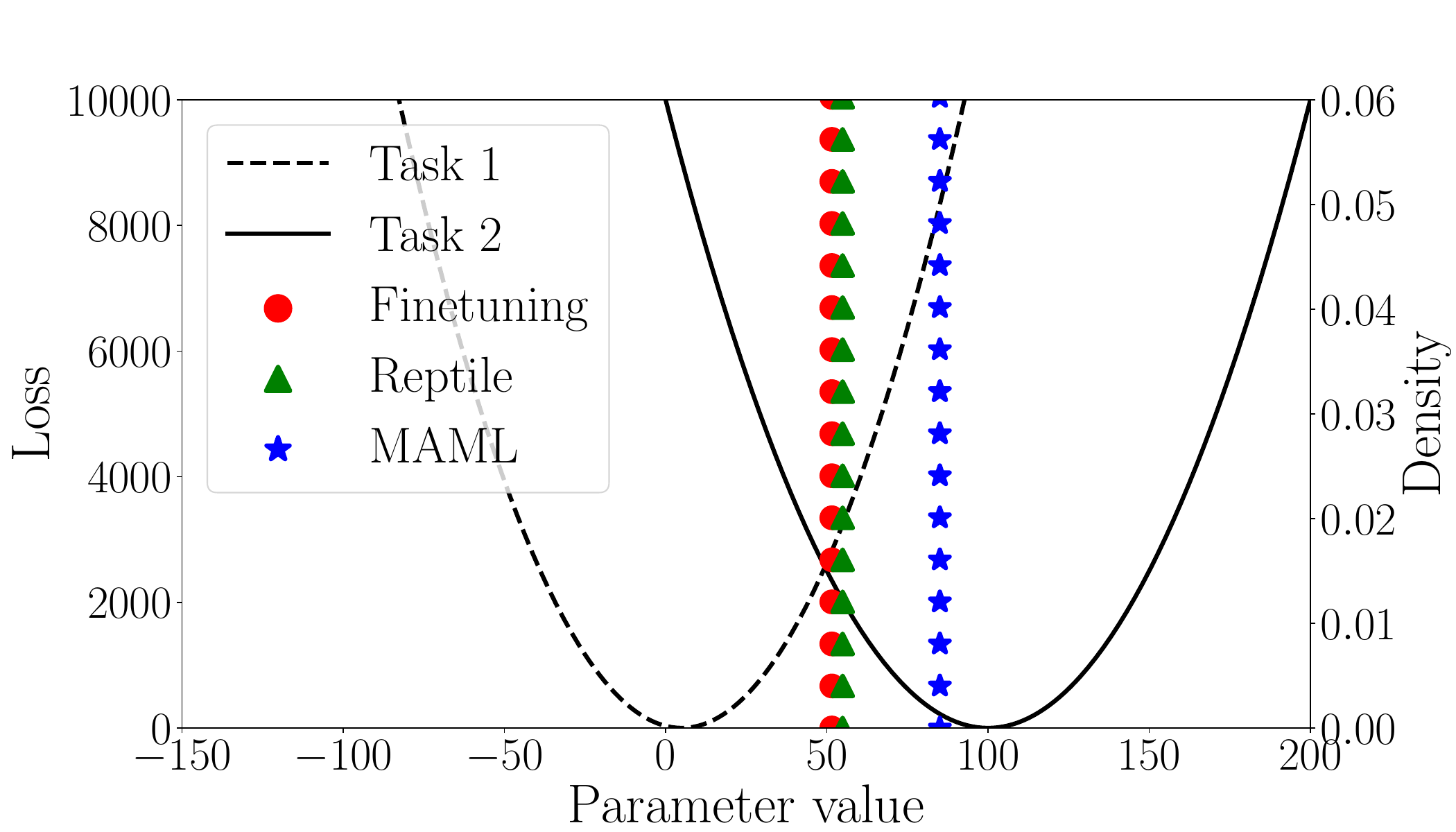}
    \caption{Scenario $a$, with $T=25$}
    \end{subfigure}
    \begin{subfigure}[b]{0.49\linewidth}
    \includegraphics[width=\linewidth]{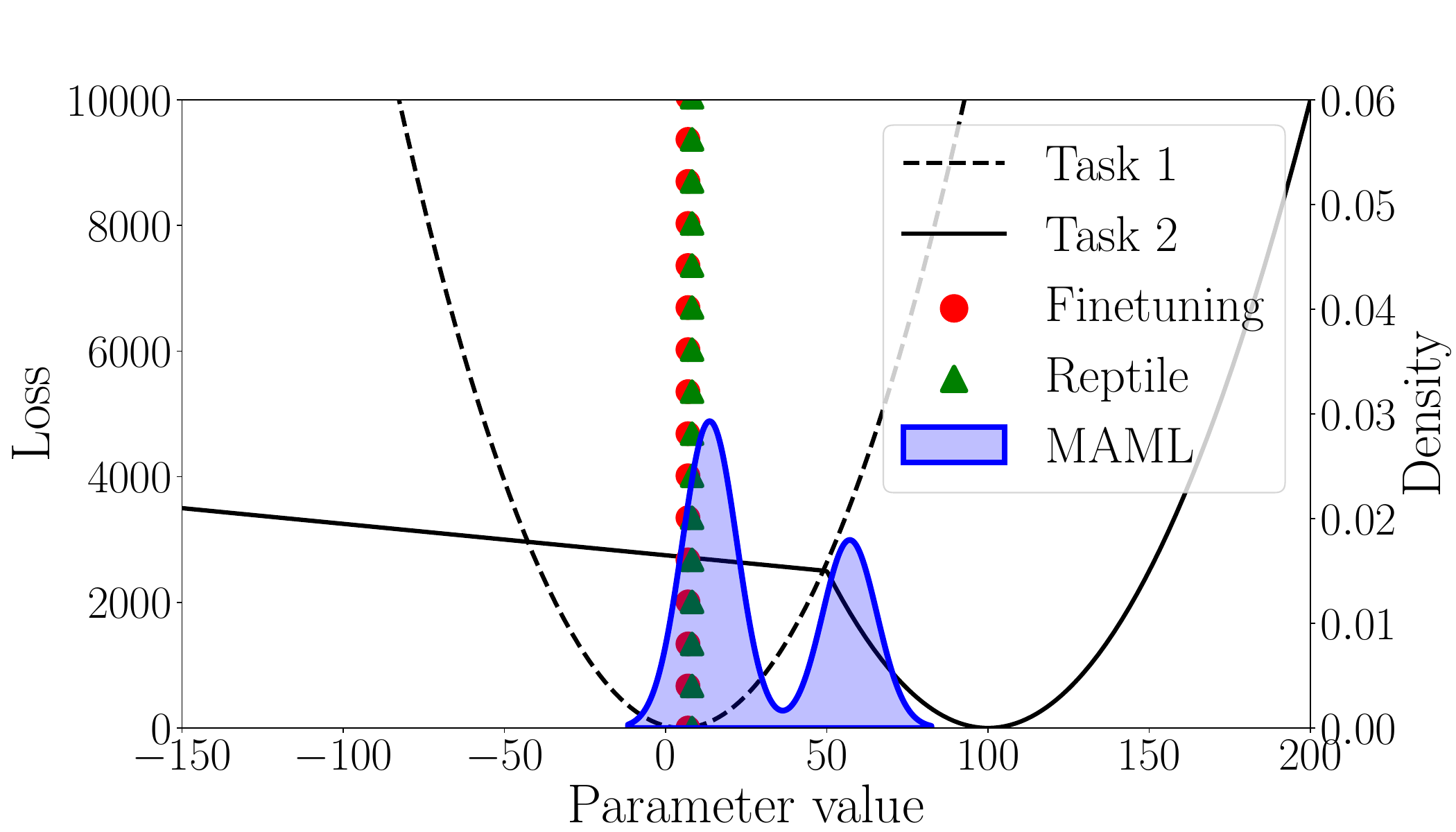}
    \caption{Scenario $b$, with $T=5$}
    \end{subfigure}
    \begin{subfigure}[b]{0.49\linewidth}
    \includegraphics[width=\linewidth]{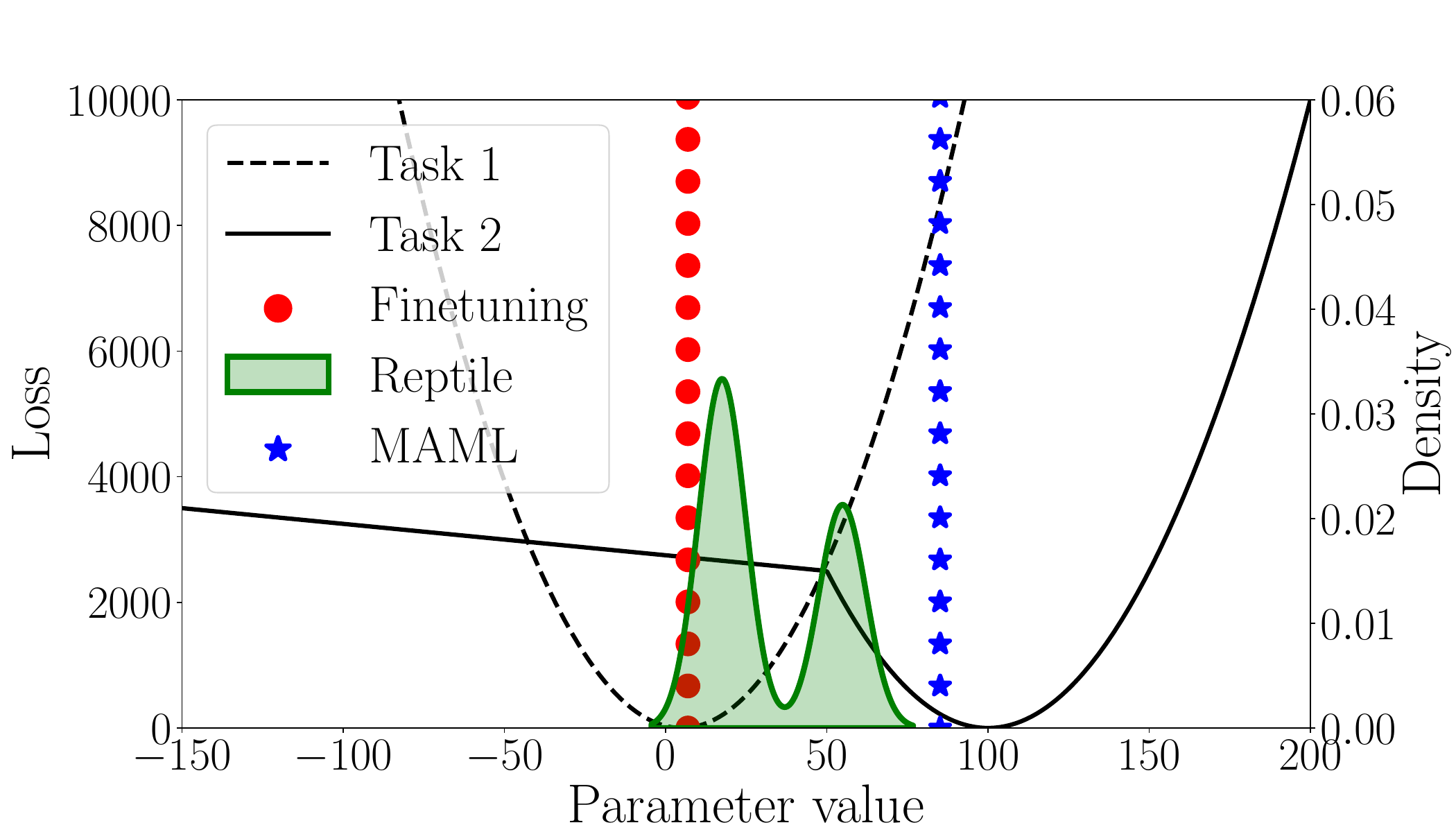}
    \caption{Scenario $b$, with $T=25$}
    \end{subfigure}
    \caption{Average initialization that finetuning, Reptile, and MAML converge to when using $T=5$ or $T=25$ adaptation steps per task. 
    In scenario $a$ (top figures), finetuning and Reptile both pick an initialization in the centre of the two optima where the initial loss is minimal. 
    MAML neglects the initial performance and thus is freer to select an initialization point, especially when $T$ is larger. 
    In scenario $b$ (bottom figures) the loss of task 2 is no longer convex and has a reasonably flat plateau. Finetuning and Reptile get stuck in the optimum of the first task and fail to learn the second task successfully, while MAML finds a location from which it can arrive at both optima.}
    \label{fig:synth}
\end{figure}

\subsection{Toy problem}\label{sec:toy}
First, we study the behaviour of finetuning, Reptile, and MAML in two synthetic scenarios $a$ and $b$, consisting of two tasks each. 
In this subsection, we use a slightly more abstract notion of tasks compared with the rest of the text, and define tasks purely abstractly by loss functions. 
These tasks can be considered the meta-train set, and the goal of the algorithms is to find good initialization parameters on this task distribution. 
We represent tasks by their loss landscape, which we have constructed by hand for illustrative purposes. 
In scenario $a$, the two task loss landscapes are quadratic functions of a single parameter $x$.
More specifically, the losses for this scenario are given by $\ell_1^a(x) = 1.3(x-5)^2$ and $\ell_2^a(x) = (x-100)^2$.
In scenario $b$, the first task loss landscape is the same $\ell^b_1 = \ell_1^a$ while the second task represents a more complex function:
\begin{align}
    \ell_2^b(x) = \begin{cases}
    (x-100)^2 & x > 50 \\
    -5x + 2750 & x \leq 50
    \end{cases}
\end{align}
The respective algorithms train by sampling tasks in an interleaved fashion, and by adapting the parameter $x$ based on the loss landscape of the sampled task.  
We investigate the behaviour of Reptile and MAML when they make $T=5$ or $T=25$ task-specific adaptation steps. 
For this, we average the found solutions of the techniques over $100$ different runs with initial $x$ values that are equally spaced in the interval $[-200, +200]$.   
We find that finetuning converges to the same point regardless of the initialization and is thus represented by a single vertical line.
For Reptile and MAML, the found solution depends on the initialization, which is why we represent the found solution as a probability density.  
A Jupyter notebook for reproducing these results can be found on our GitHub page.

Based on the learning objectives of the techniques, we expect finetuning to settle for an initialization that has a good initial performance on both tasks (small loss values).
Furthermore, we expect that MAML will pick any initialization point from which it can reach minimal loss on both tasks within $T$ steps. 
Reptile is expected to find a mid-way solution between finetuning and MAML. 

The results of these experiments are displayed in \autoref{fig:synth}. 
In scenario $a$ (top figures), we see that both finetuning and Reptile prefer an initialization at the intersection of the two loss curves, where the initial loss is minimal. 
MAML, on the other hand, neglects the initial performance when $T=25$ and leans more to the right, whilst ensuring that it can reach the two optima within $T$ steps. 
The reason that it prefers an initialization on the right of the intersection is that the loss landscape of task 1 is steeper, which means that task adaptation steps will be larger. 
Thus, a location at the right of the intersection ensures good learning of task 2 and yields comparatively fast learning on the first task.

In scenario $b$ (bottom figures), the loss landscape of task 2 has a relatively flat plateau on the left-hand side. 
Because of this, finetuning and Reptile will be pulled towards the optimum (also the joint optimum) of the first task due to the larger gradients compared with the small gradients of the flat region of the second task when $T$ is small.
The solution that is found by MAML when $T=5$ depends on the random initialization of the parameter, as can be seen in plot c). 
That is, when the random initialization is on the left of the plateau, MAML can not look beyond the flat region, implying that it will also be pulled towards the minimum of task 1. 
When $T=25$, allowing the Reptile and MAML to look beyond the flat region, we see that Reptile either finds an initialization at $x=50$ (when the starting point $x_0$ is on the right-hand side of the plateau) or at the joint optimum at $x=0$ (when it starts with $x_0$ on the plateau).
In the latter case, the post-adaptation performance of Reptile on both tasks is not optimal because it cannot reach the optimum of task 2.
MAML, on the other hand, does not suffer from this suboptimality because it neglects the initial and intermediate performance and simply finds an initialization at $x \approx 85$ from which it can reach both the optima of tasks 1 and 2.

\subsection{Few-shot image classification}

We continue our investigations by studying why MAML and Reptile can outperform finetuning in within-distribution few-shot image classification settings (see \autoref{sec:fsl}) when using a Conv-4 backbone.
For these experiments, we use the $N$-way $k$-shot classification setting (see \autoref{sec:fsl}) on the miniImageNet \citep{vinyals2016matching,ravi2017optimization} and CUB \citep{wah2011caltech} benchmarks. 
miniImageNet is a mini variant of the large ImageNet dataset \citep{deng2009imagenet} for image classification, consisting of $60\,000$ colored images of size $84 \times 84$. The dataset contains 100 classes and 600 examples per class. We use the same train/validation/test class splits as in \citet{ravi2017optimization}. 
The CUB dataset contains roughly $12\,000$ RGB images of birds from 200 species (classes). We use the same setting and train/validation/test class splits as in \citet{chen2019closer}.

Note that using real datasets entails that we move away from the abstract task definition as in the previous toy experiment, where the loss signal of the task was perfect.
Instead, the loss signal is now approximated by sampling a finite set of data points for every task (for MAML and Reptile) or batch (for finetuning) and computing the performance of the methods on it. 

For finetuning and MAML, we tune the hyperparameters on the meta-validation tasks using random search with a budget of 30 function evaluations for every backbone and dataset. 
We train MAML on $60\,000$ tasks in the 1-shot setting and on $40\,000$ tasks in the 5-shot setting, and validate its performance every $2\,500$ tasks.
The checkpoint with the highest validation accuracy is then evaluated on $600$ holdout test tasks.
Similarly, finetuning is trained on $60\,000$ batches of data from the training split when we evaluate it in the 1-shot setting and on $40\,000$ batches when evaluating it in the 5-shot setting. 
Note that finetuning is trained on simple mini-batches of data instead of tasks consisting of a support and query set, and is later validated and tested on unseen validation and test tasks, respectively. 
In a similar fashion as for MAML, we validate its performance every $2\,500$ batches. 
Due to the computational expenses, for Reptile, we use the best-reported hyperparameters and training iterations on 5-way 1-shot miniImageNet as found by \citet{nichol2018reptile}.
We use Torchmeta for the implementation of the data loaders \citep{deleu2019torchmeta}.
We note that a single run of MAML and finetuning finish within one day, while Reptile finished within 4 days, perhaps due to the absence of parallelism in the implementation we used.

\subsubsection{The role of the output layer}\label{sec:output}

\begin{figure}
    \centering
    \begin{subfigure}[]{.48\textwidth}
    \includegraphics[width=\linewidth]{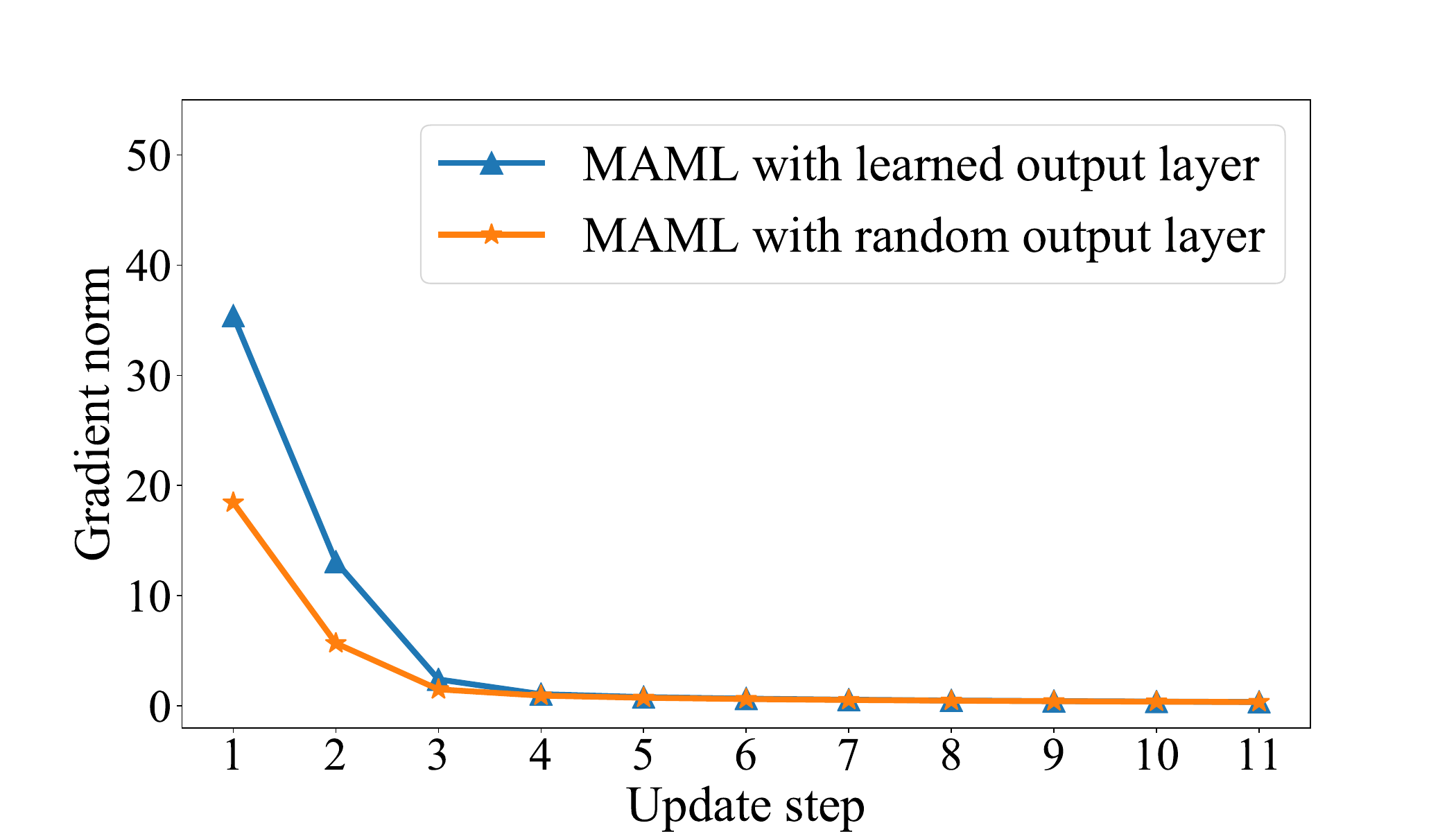}
    \caption{MAML on MIN}
    \end{subfigure}
    \begin{subfigure}[]{.48\textwidth}
    \includegraphics[width=\linewidth]{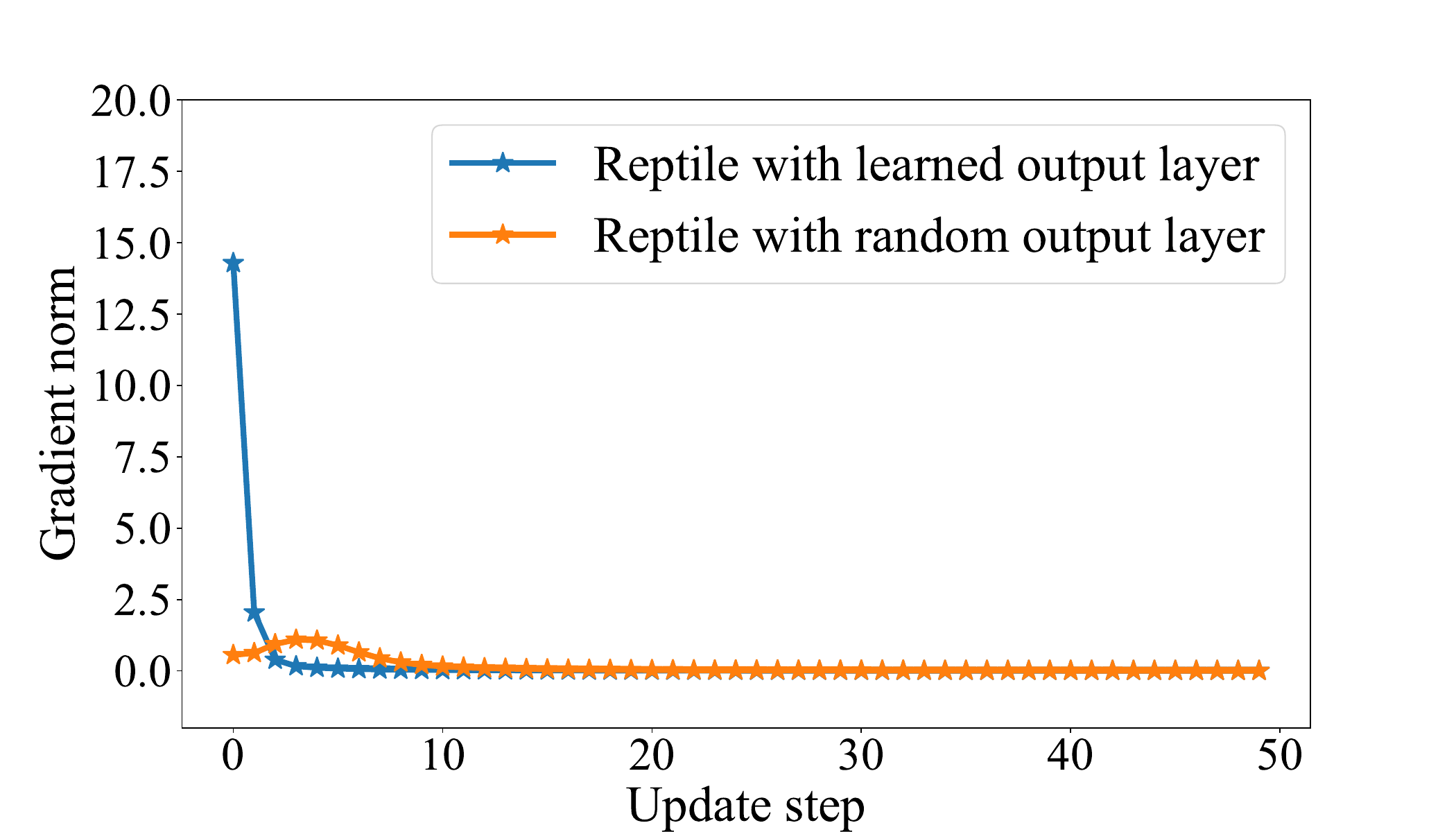}
    \caption{Reptile on MIN}
    \end{subfigure}
    \begin{subfigure}[]{.48\textwidth}
    \includegraphics[width=\linewidth]{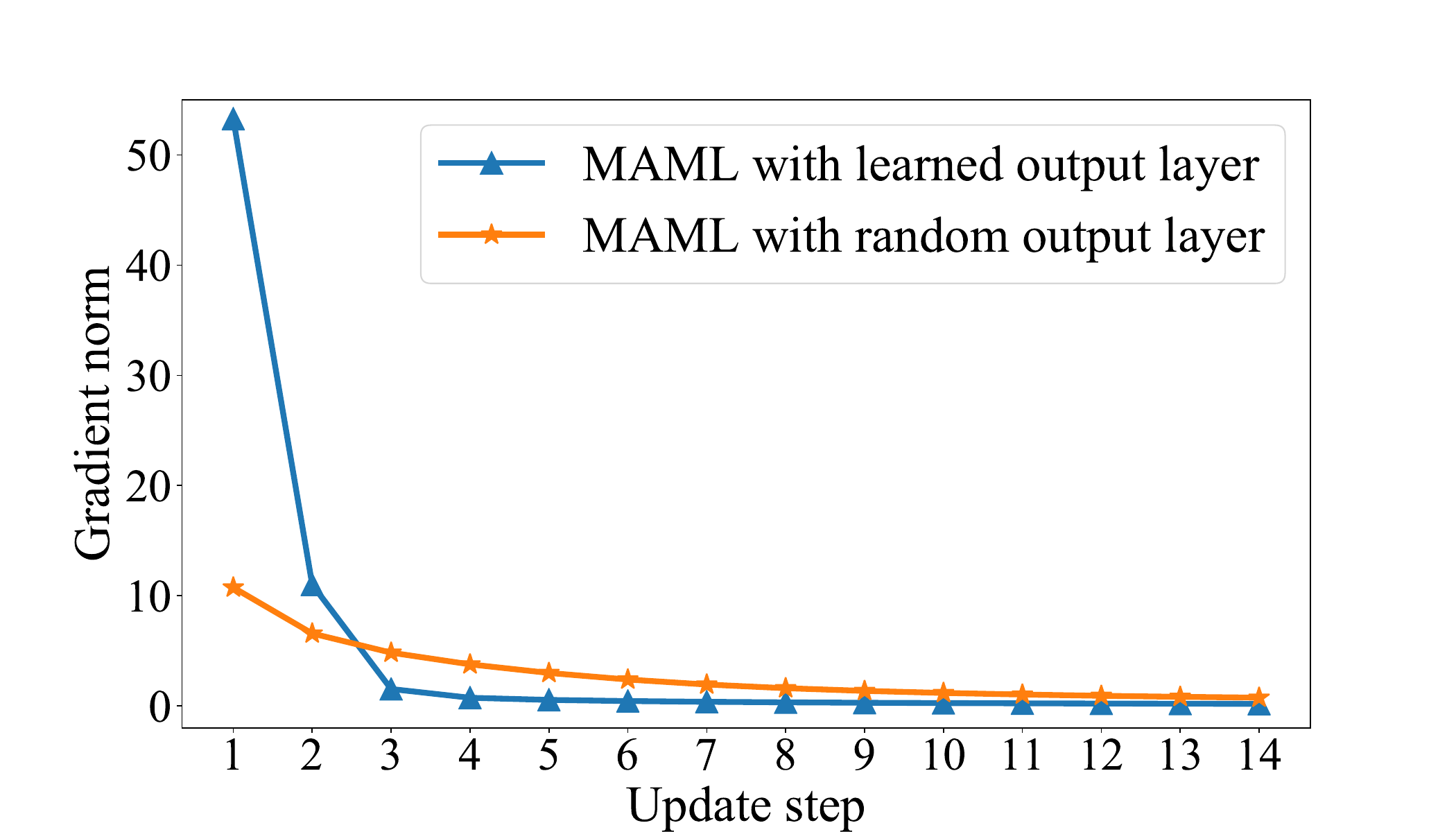}
    \caption{MAML on CUB}
    \end{subfigure}
    \begin{subfigure}[]{.48\textwidth}
    \includegraphics[width=\linewidth]{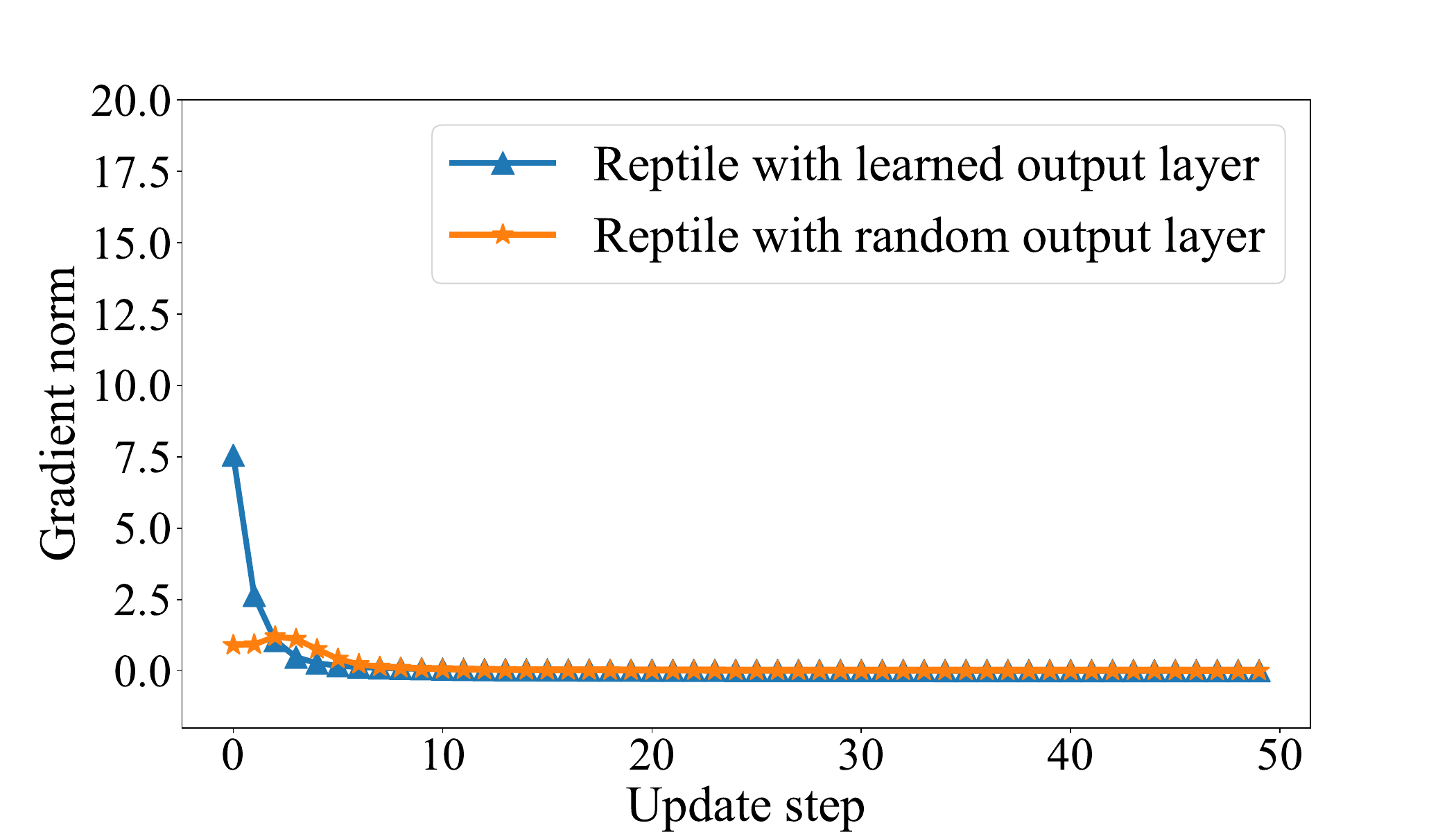}
    \caption{Reptile CUB}
    \end{subfigure}
    \caption{The difference in the average gradient norms during inner-loop adaptation between MAML (left) and Reptile (right) with a learned output layer and a randomly initialized one on 5-way 1-shot miniImageNet (MIN; top row) and CUB (bottom row). The 95\% confidence intervals are within the size of the symbols. The learned output layers have a higher gradient norm at the beginning of the training phase.}
    \label{fig:randoutgrads}
\end{figure}

Here, we investigate whether the fact that MAML and Reptile reuse their learned output layer when learning new tasks alter their inner-learning behaviour and give them an advantage in performance compared with finetuning, which removes the learned output layer and replaces it with a randomly initialized one when learning a new task. 
In short, we study the role of the output layer on the performance and inner-loop adaptation behaviour of MAML and Reptile.
For this, we perform meta-training for MAML and Reptile on 5-way 1-shot miniImageNet classification, and study the effect of replacing the learned output layer initialization weights with random weights on their ability to learn new tasks.  
Note that even though the weight initialization of the output layer may be random, it is still trained on the support sets of unseen tasks, therefore, finetuned to the task upon which it will be evaluated. 
\autoref{fig:randoutgrads} displays the effect of replacing the output layer of the meta-learned weight initialization by MAML and Reptile meta-trained on 5-way 1-shot miniImageNet, with a randomly initialized one on the gradient norms during the inner-loop adaptation procedure.
As we can see, the networks of the variants with a learned output layer receive larger gradient norms at the first few updates compared with the variants using a randomly initialized output layer, indicating that the learned output layer alters the learning behaviour of the algorithms.
However, at the end of adaptation for a given task, the gradient norms are close to zero for both variants, indicating that both have converged to a local minimum. 
This implies that the learned initialization of the output layer has a distinct influence on the learning behaviour of new tasks.
More specifically, using a learned output layer may aid in finding an initialization in the loss landscape that is sensitive to tasks and can be quickly adapted, explaining the larger gradient norms. 

\begin{figure}
    \centering
    \begin{subfigure}[]{.48\textwidth}
    \includegraphics[width=\linewidth]{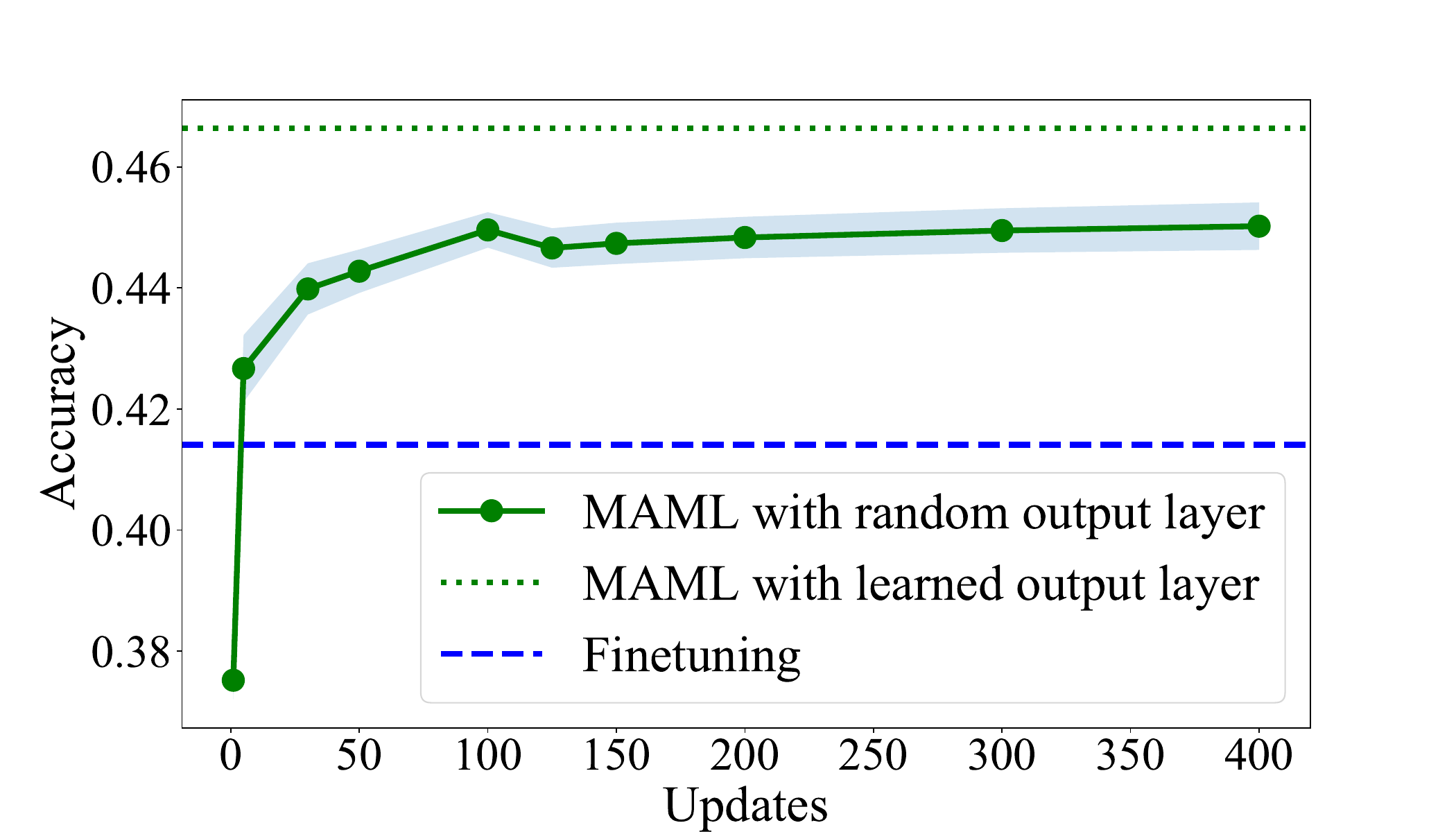}
    \caption{MAML on MIN}
    \end{subfigure}
    \begin{subfigure}[]{.48\textwidth}
    \includegraphics[width=\linewidth]{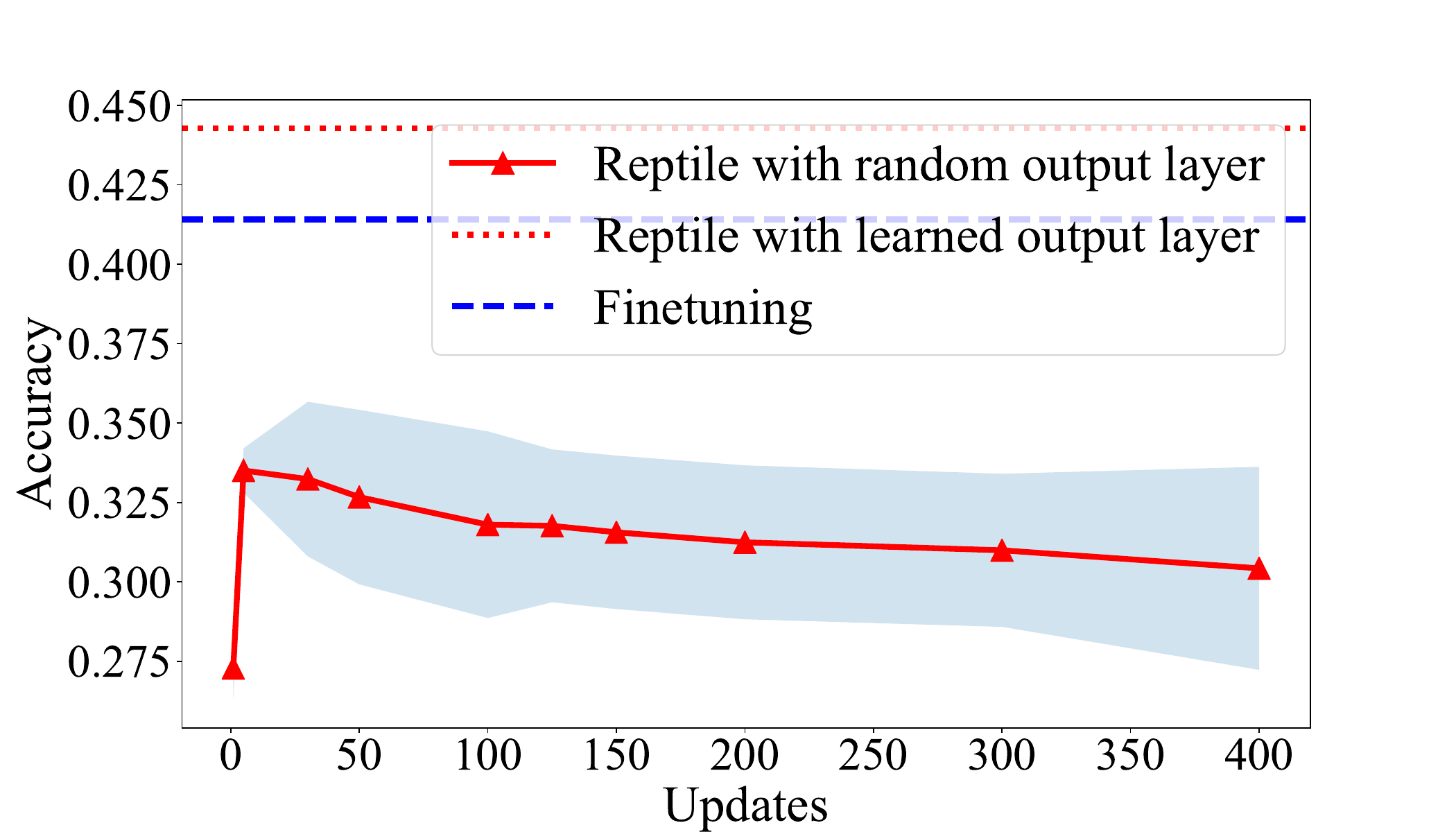}
    \caption{Reptile on MIN}
    \end{subfigure}
    \begin{subfigure}[]{.48\textwidth}
    \includegraphics[width=\linewidth]{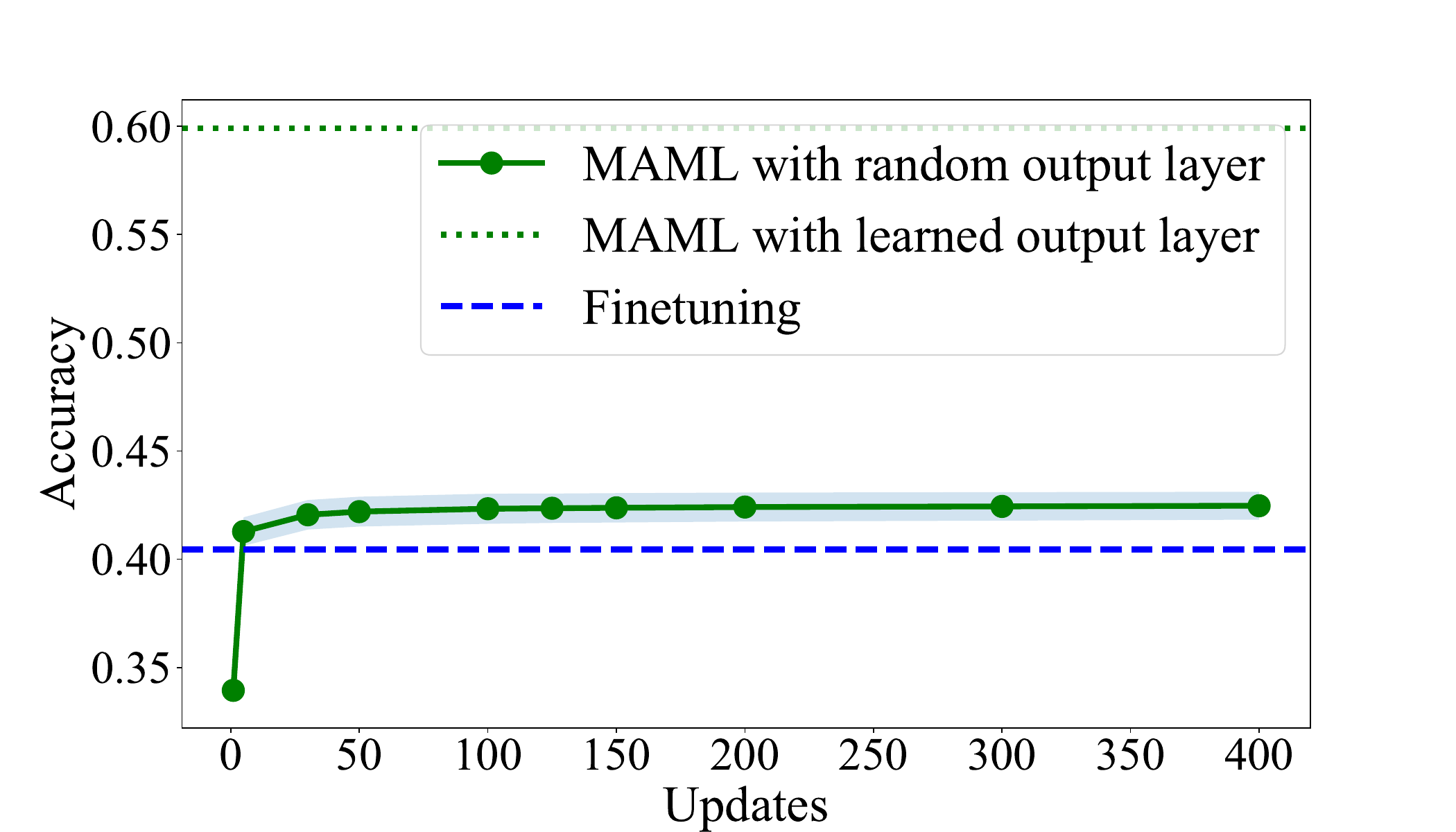}
    \caption{MAML on CUB}
    \end{subfigure}
    \begin{subfigure}[]{.48\textwidth}
    \includegraphics[width=\linewidth]{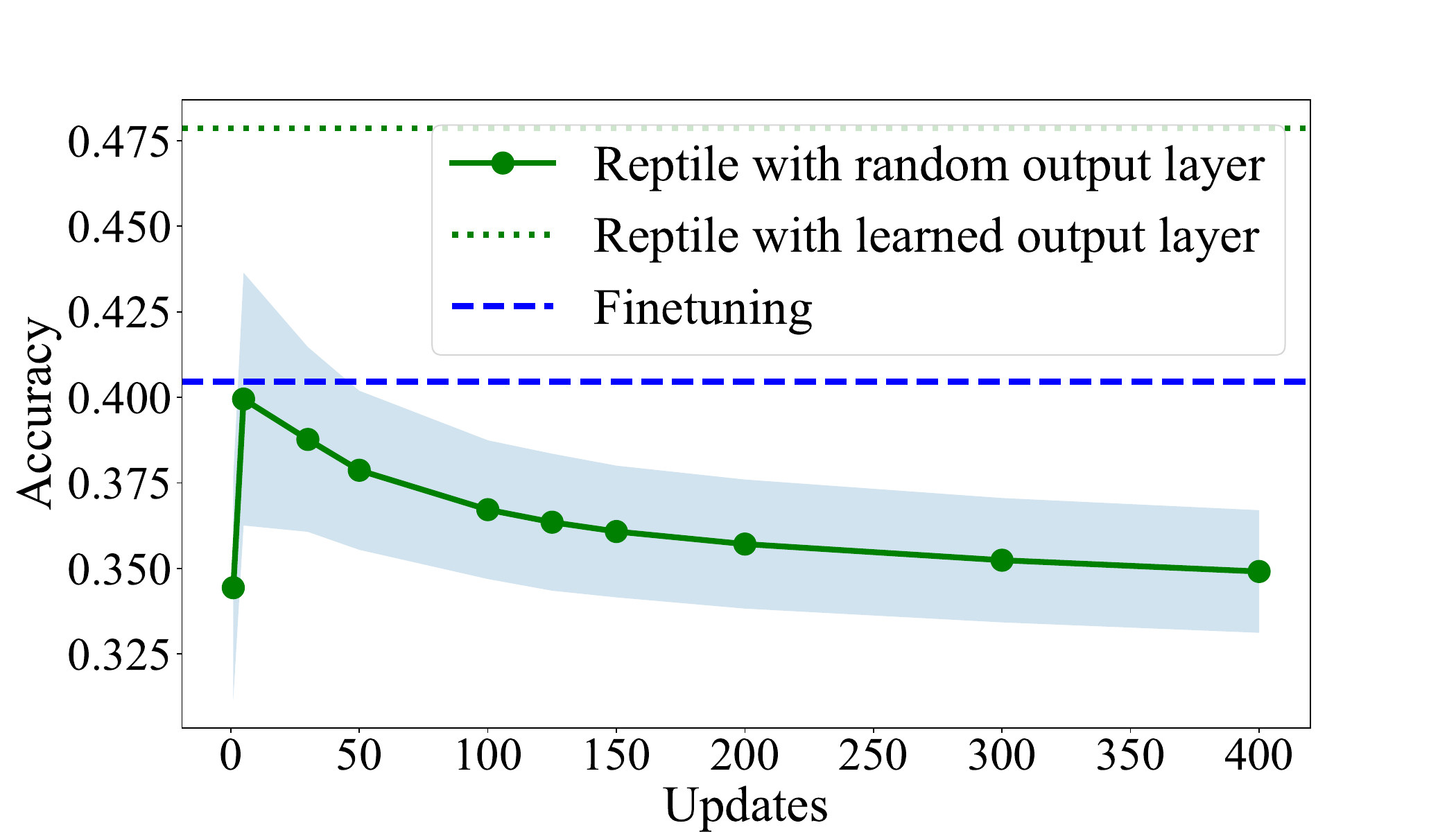}
    \caption{Reptile on CUB}
    \end{subfigure}
    \caption{The difference in performance between MAML (left) and Reptile (right) with a learned output layer and a randomly initialized one on 5-way 1-shot miniImageNet (MIN; top row) and CUB (bottom row) for different numbers of update steps. The 95\% confidence intervals are displayed as shaded regions. Learning new tasks starting with a random output layer fails to achieve the same end performance as with the learned output layer.}
    \label{fig:randout}
\end{figure}

Next, we investigate whether reusing the learned output layers also leads to performance differences. 
For this, we investigate the influence of replacing the learned output layers in MAML and Reptile with randomly initialized ones when starting to learn new tasks on their learning performance for different numbers of update steps.
The results are shown in \autoref{fig:randout}. 
As we can see, replacing the output layer with a random one leads to worse performance.
Increasing the number of updates improves the performance for MAML, while the reverse is true for Reptile.  
In the end, the performance gap introduced by replacing the output layers with random ones is not closed, indicating that the output layers play an important role in successful inner-loop adaptation.

\subsubsection{Specialization for robustness against overfitting}\label{sec:noise}

In this subsection, we investigate the influence of the level of data scarcity in the support set on the performance of MAML and Reptile.
We hypothesize that both algorithms learn an initialization that is robust against overfitting when the number of examples in the support set per class ($k$) is small. 
This would imply that their performance would suffer when the number of examples in the support sets in training tasks is large due to the reduced need to become robust against overfitting, disabling the meta-learning techniques to become robust to overfitting during task-specific adaptation.
We investigate this for 5-way miniImageNet image classification by varying the number of examples in the support set of meta-training tasks and measuring the performance on tasks with only one example per class (1-shot setting). 

\begin{figure}
    \centering
    \begin{subfigure}[]{.48\textwidth}
    \includegraphics[width=\linewidth]{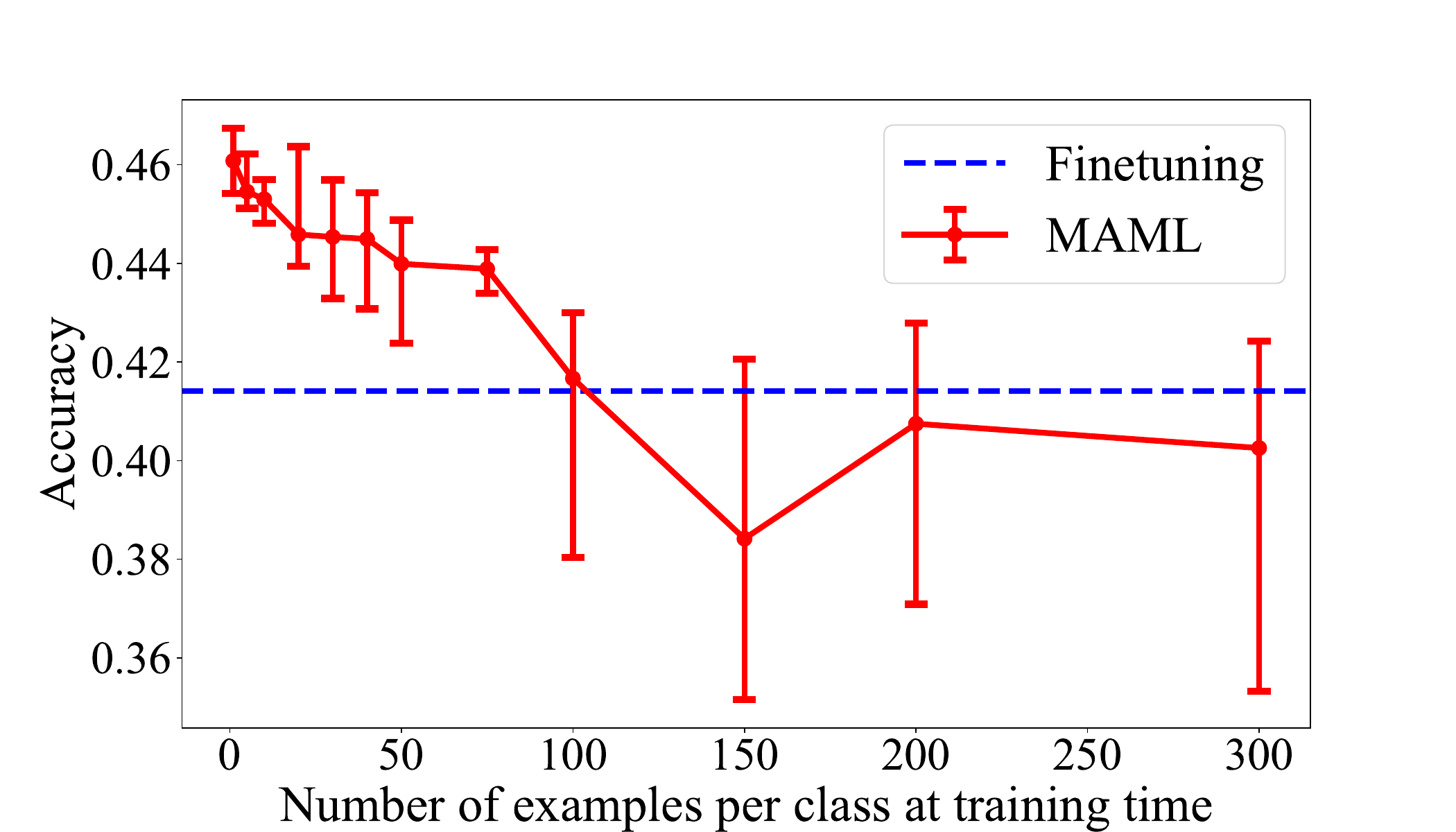}
    \caption{MAML on MIN}
    \end{subfigure}
    \begin{subfigure}[]{.48\textwidth}
    \includegraphics[width=\linewidth]{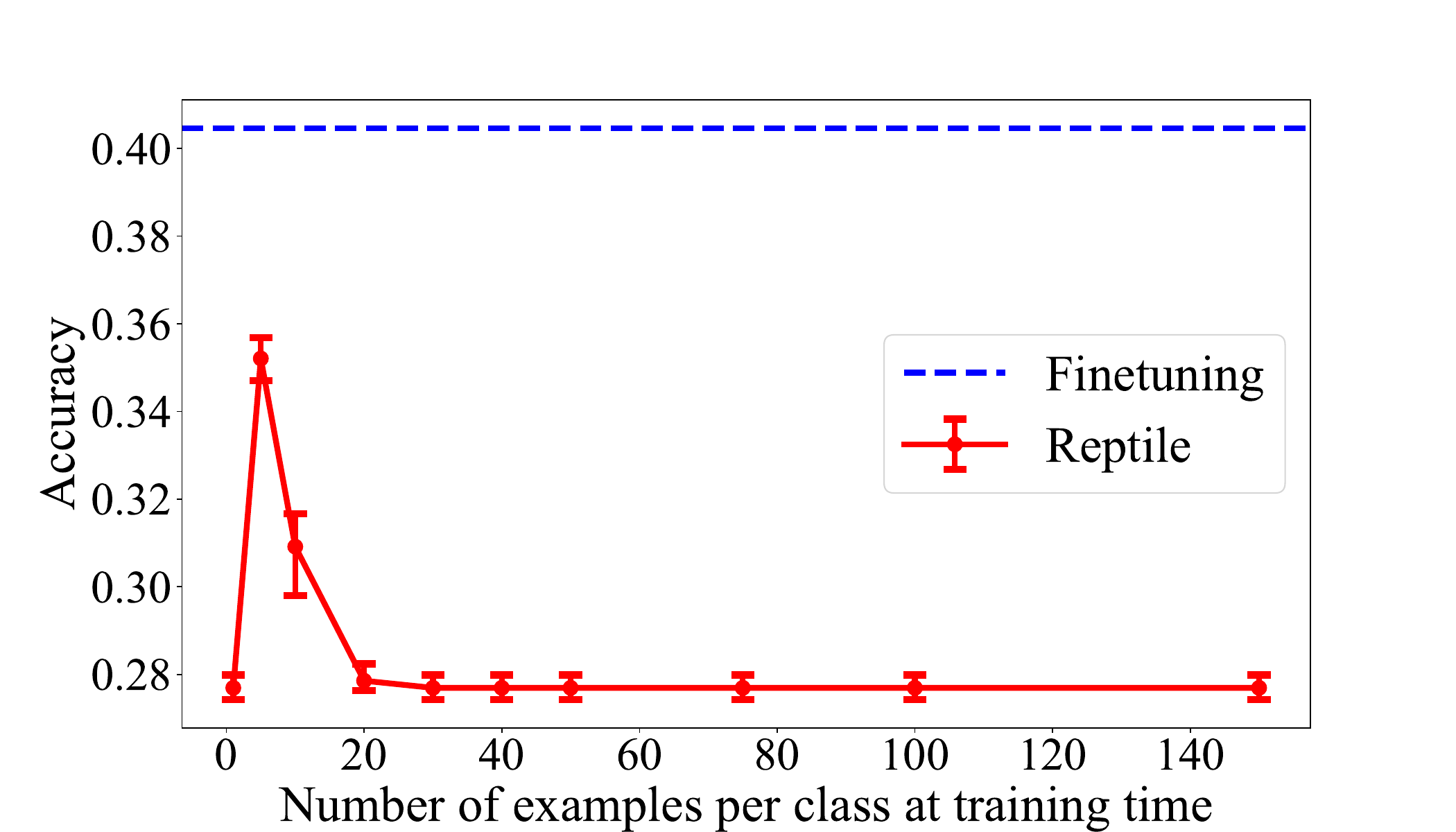}
    \caption{Reptile on MIN}
    \end{subfigure}
    \begin{subfigure}[]{.48\textwidth}
    \includegraphics[width=\linewidth]{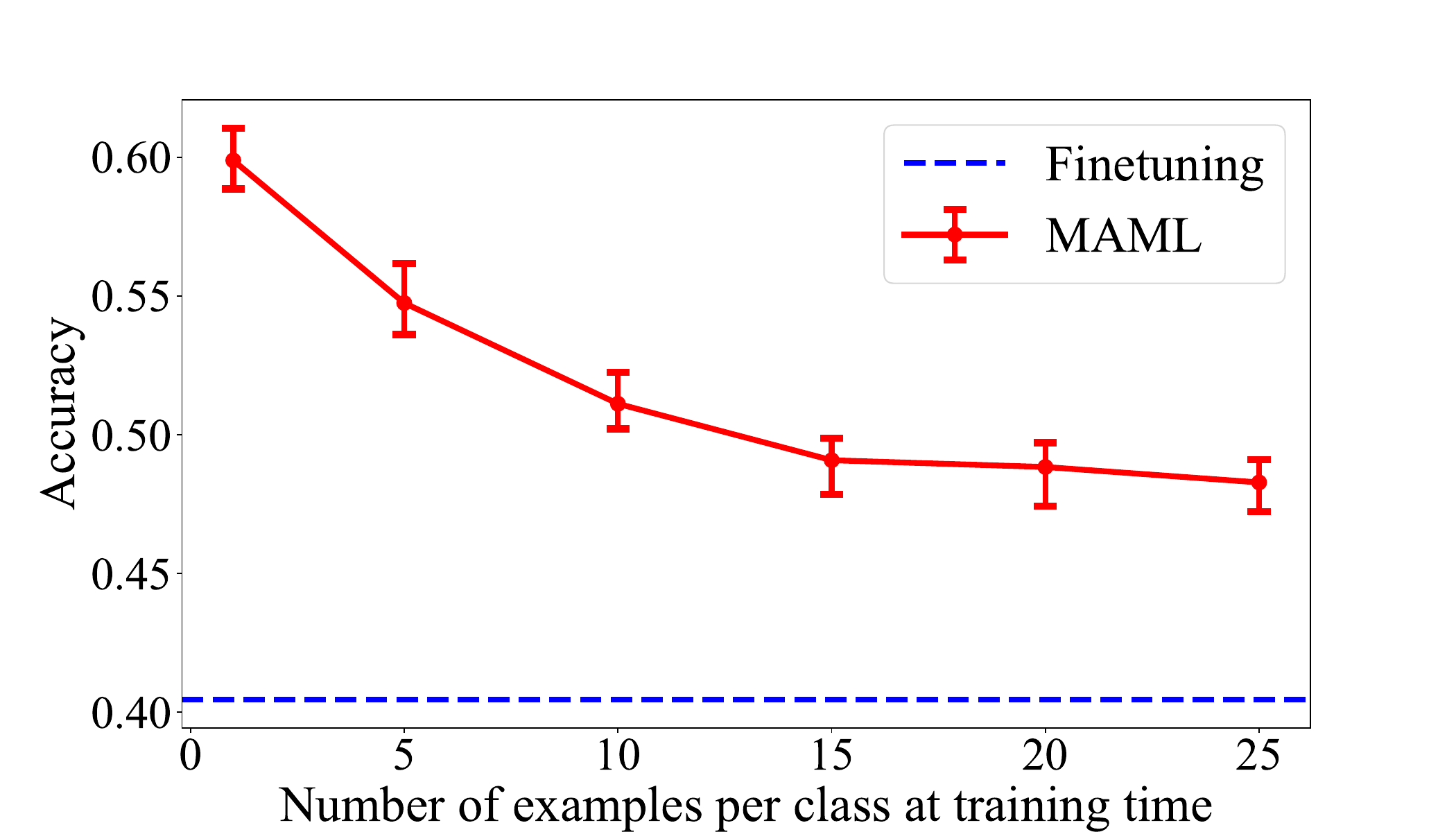}
    \caption{MAML on CUB}
    \end{subfigure}
    \begin{subfigure}[]{.48\textwidth}
    \includegraphics[width=\linewidth]{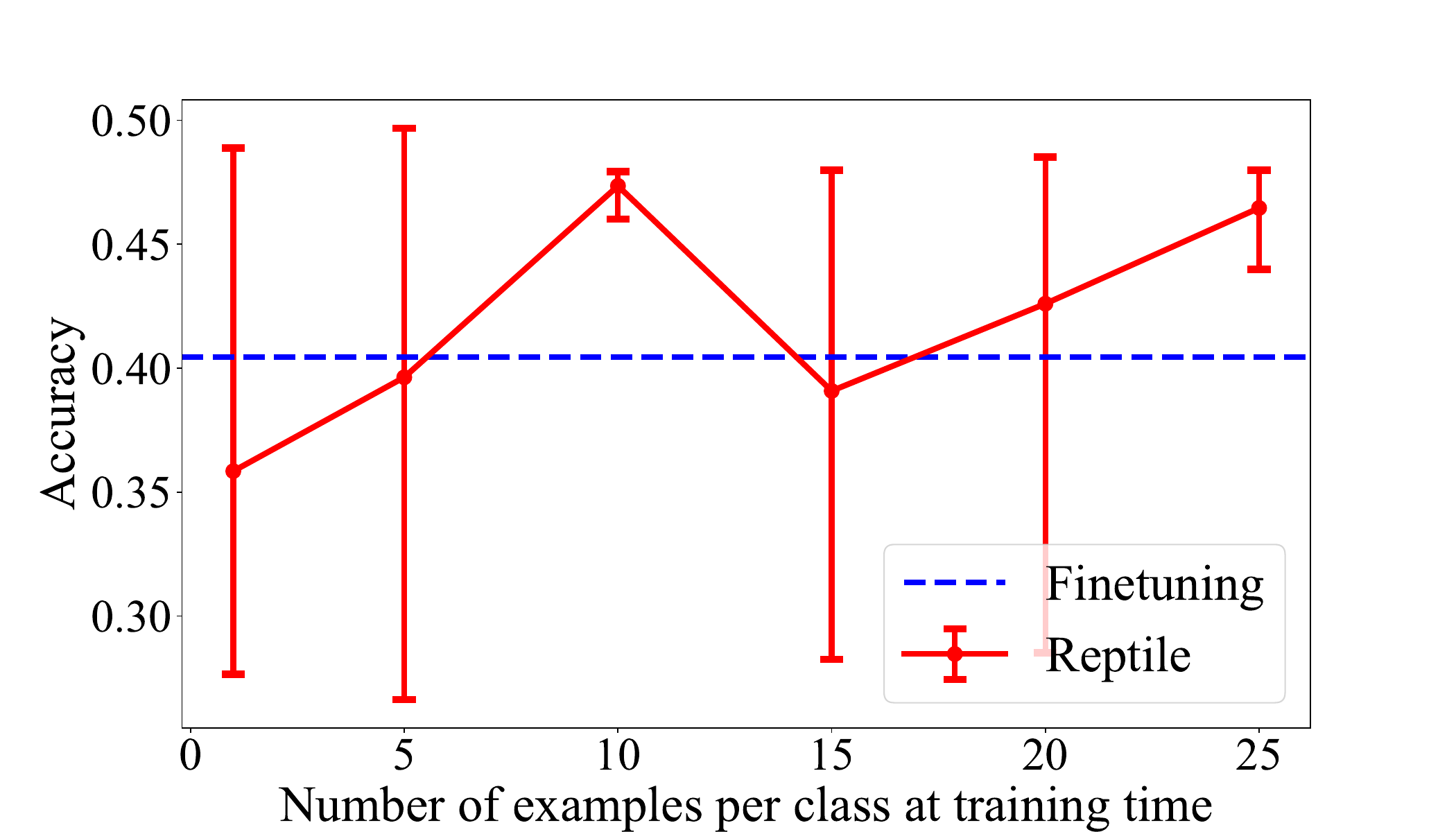}
    \caption{Reptile on CUB}
    \end{subfigure}
    \caption{The effect of the number of training examples per class in the support set on the performance of MAML (left) and Reptile (right) on 5-way 1-shot miniImageNet (MIN; top row) and CUB (bottom row) classification. The larger the number of examples, the worse the few-shot learning performance of MAML. The error bars show the maximum and minimum performance over 5 runs with different random seeds. Note that the test tasks contain only a single example per class in the support set.
    }
    \label{fig:noise}
\end{figure}

\autoref{fig:noise} displays the results of these experiments. 
As we can see, there is an adverse effect of increasing the number of support examples per task on the final 1-shot performance of MAML. 
This shows that for MAML, it is important to match the training and test conditions so the initialization parameters can become robust against overfitting induced by data scarcity.
In addition, we observe that Reptile is unstable due to its sensitivity to different hyperparameters on miniImageNet, even in the setting where $k=1$. This is caused by the fact that Reptile is not allowed to sample mini-batches of data from the support set. Instead, we force it to use the full support set to investigate the effect of the number of support examples. 
When the number of examples is close to ten, which is the mini-batch size commonly used, as by the original authors \citep{nichol2018reptile}, there is a slight increase in performance for Reptile on miniImageNet, supporting the observation that it is sensitive to the chosen hyperparameters.  
On CUB, in contrast, we observe that the performance improves with the number of examples per class at training time, although the maximum number of examples investigated is 25 due to the fact that not every class has more examples than that.
This illustrates that the sensitivity to hyperparameters depends on the chosen dataset.

\begin{figure}[tb]
    \centering
    \includegraphics[width=0.8\linewidth, height=150px]{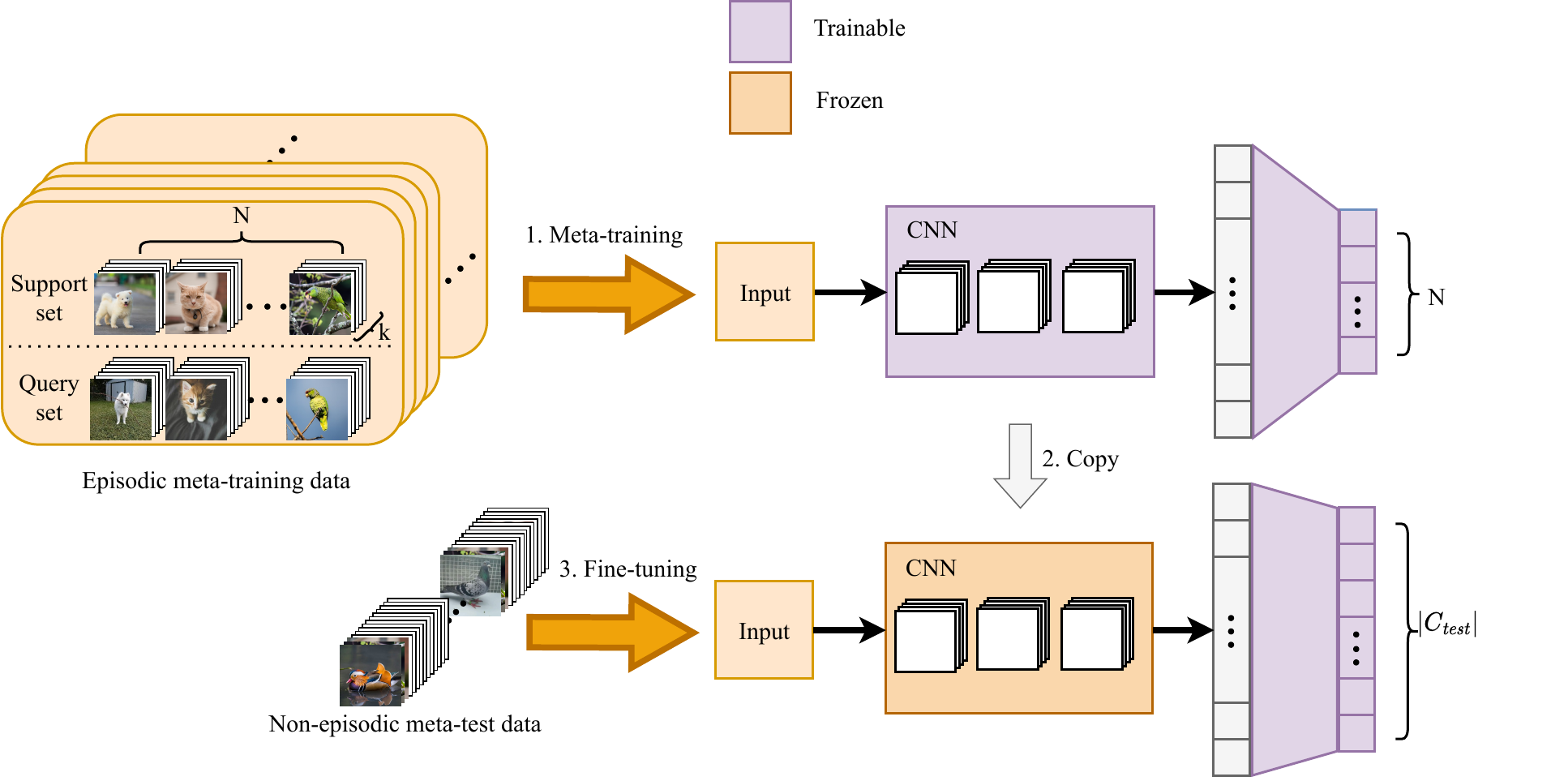}
    \caption{Flow chart for measuring the joint classification accuracy for meta-learning techniques. 
    First, we train the techniques in an episodic manner on all data in the meta-train set. 
    Second, we copy and freeze the learned initialization parameters and replace the output layer with a new one. 
    Third, we fine-tune this new output layer on all meta-test data in a non-episodic manner. 
    As such, the meta-test data is split into a non-episodic train and a non-episodic test set. 
    Finally, we evaluate the learned evaluation on the hold-out test split of the meta-test data.
    We refer to the resulting performance measure as the joint classification accuracy. 
    Note that finetuning follows the same procedure, with the exception that it trains non-episodically (on batches instead of tasks) on the meta-training data.}
    \label{fig:jointacc}
\end{figure}

\subsubsection{Information content in the learned initializations}\label{sec:infcont}

Next, we investigate the relationship between the few-shot image classification performance and the discriminative power of the learned features by the three techniques for different backbones (Conv-4, ResNet10, ResNet18 \citep{he2015delving}). 

\begin{figure}[htb]
    \centering
    \begin{subfigure}[]{.48\textwidth}
    \includegraphics[width=\linewidth]{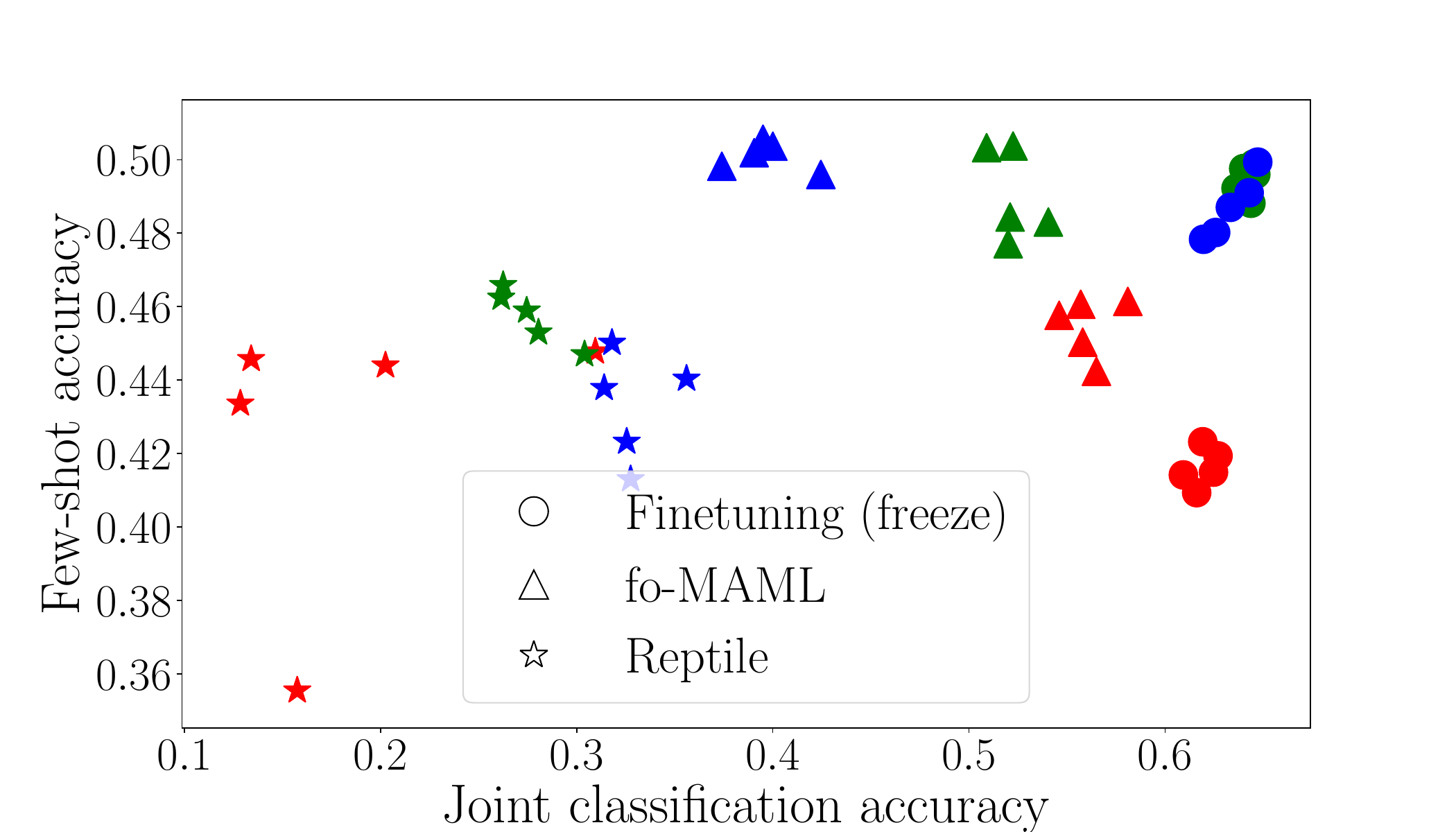}
    \caption{miniImageNet (r=0.34, p=0.02)}
    \end{subfigure}
    \begin{subfigure}[]{.48\textwidth}
    \includegraphics[width=\linewidth]{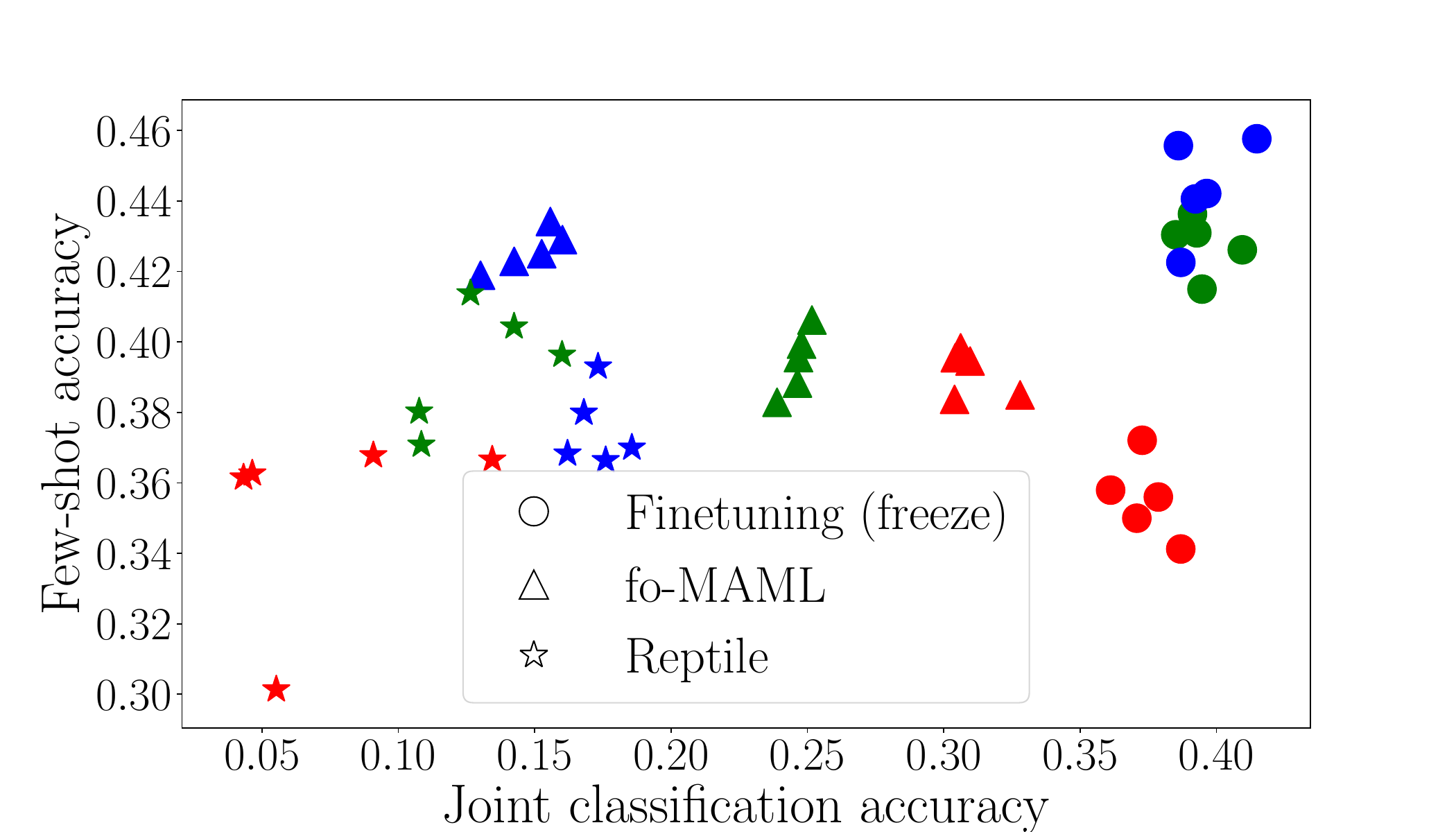}
    \caption{miniImageNet $\rightarrow$ CUB (r=0.39, p=9e-3)}
    \end{subfigure}
    \begin{subfigure}[]{.48\textwidth}
    \includegraphics[width=\linewidth]{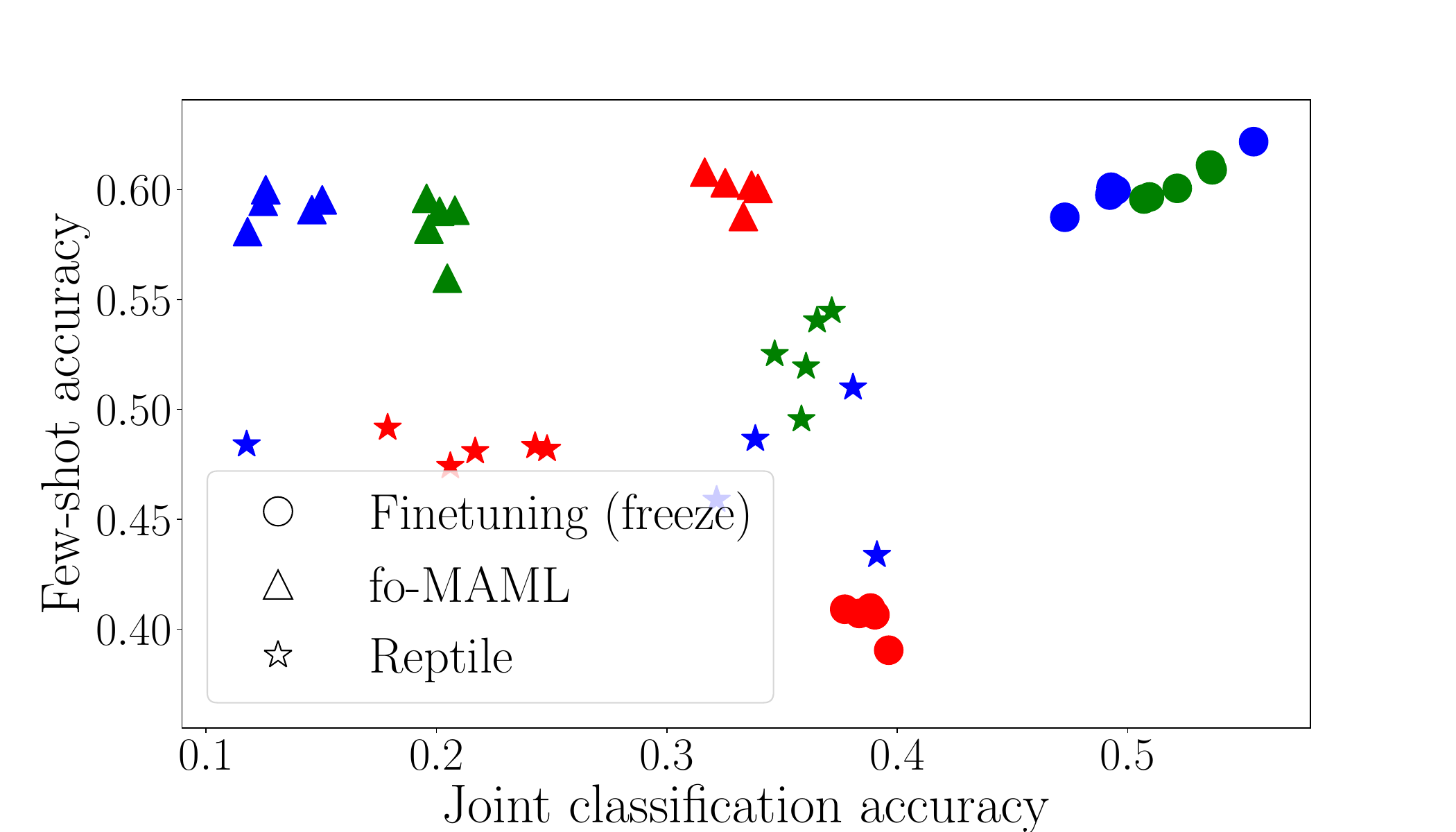}
    \caption{CUB (r=0.07, p=0.63)}
    \end{subfigure}
    \begin{subfigure}[]{.48\textwidth}
    \includegraphics[width=\linewidth]{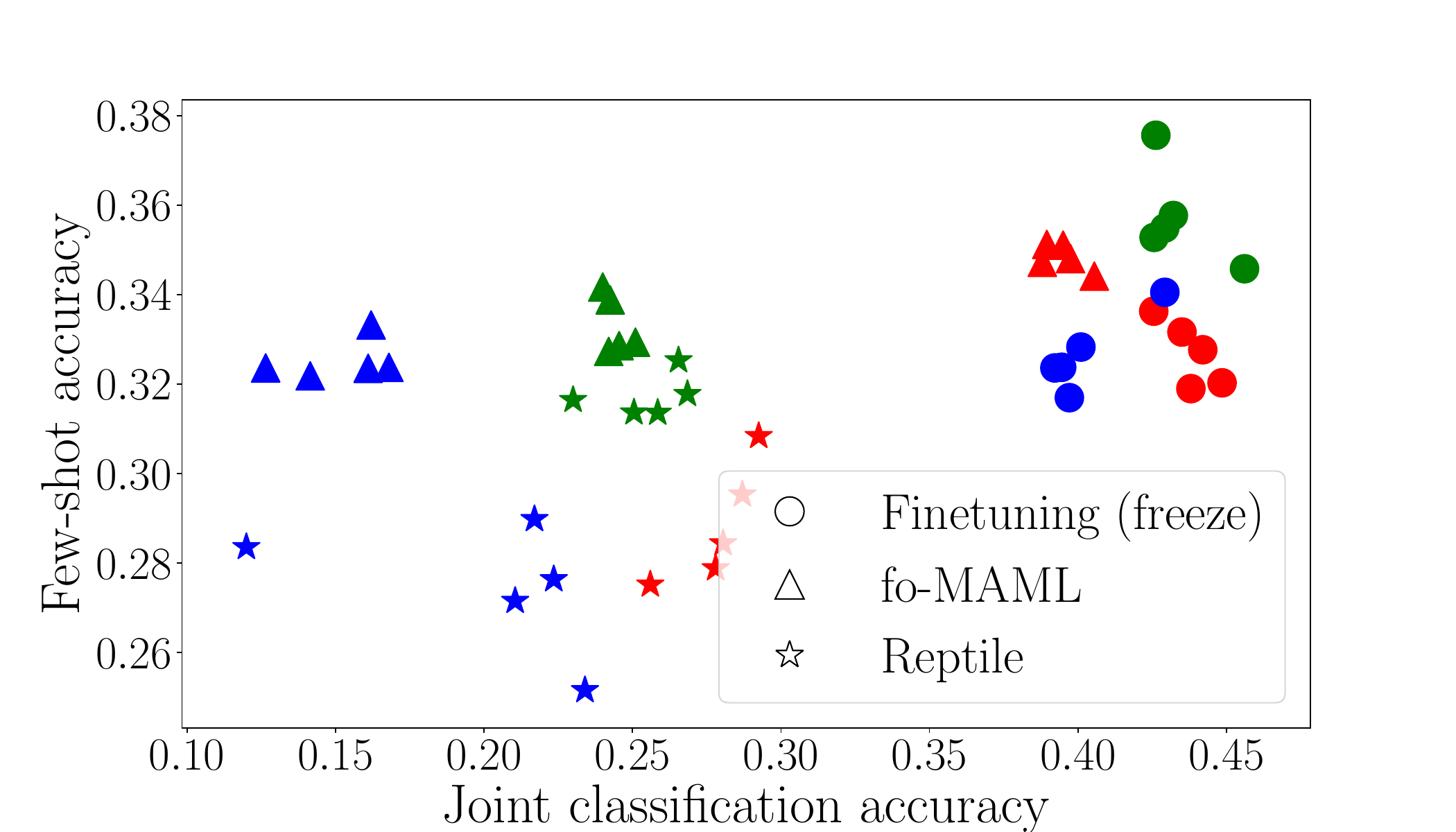}
    \caption{CUB $\rightarrow$ miniImageNet (r=0.53, p=2e-4)}
    \end{subfigure}
    \caption{The joint classification accuracy (x-axes) plotted against the $5$-way $1$-shot performance (y-axis) on all test classes. For every technique, there are 15 results plotted, corresponding to 3 backbones (Conv-4=red, ResNet-10=green, ResNet-18=blue) and 5 runs per setting. The Pearson correlation coefficients (r) and p-values are displayed in the subcaptions. The general correlations between the few-shot learning performance and joint classification accuracy range from weak to mild.}
    \label{fig:corrs}
\end{figure}

After deploying the three techniques on the datasets in a 5-way 1-shot manner, we measure the discriminative power of the learned initializations. 
\autoref{fig:jointacc} visualizes this procedure for MAML and Reptile; finetuning follows a similar procedure.
First, we extract the learned initialization parameters from the techniques. 
Second, we load these initializations into the base-learner network, freeze all hidden layers, and replace the output layer with a new one.
The new output layer contains one node for every of the $ \vert C_{test} \vert $ classes in the meta-test data.
Third, we fine-tune this new output layer on the meta-test data in a \emph{non-episodic} manner, which corresponds to regular supervised learning on the meta-test dataset.
We use a 60/40 train/test split and evaluate the final performance on the latter.   
We refer to the resulting performance measure as the \emph{joint classification accuracy}, which aims to indicate the discriminative power of the learned initialization, evaluated on data from unseen classes. 
Note that we use the expressions ``discriminative power'' and ``information content'' of the learned backbone synonymously. 

\begin{table}[tb]
\centering
\caption{Individual correlations between the joint classification accuracy and the few-shot learning performance. 
The Pearson correlation coefficients are indicated as \emph{r} and corresponding p-values as \emph{p}. 
We note that the results for each of the three few-shot learning techniques are produced with three different backbone networks. 
As such, correlations should be interpreted with utmost care. 
Significant correlations (using a threshold of $\alpha = 0.005$) are displayed in bold. 
``MIN": miniImageNet.}
\label{tab:correlations}
\begin{adjustbox}{width=\linewidth}
\begin{tabular}{lllll}
\toprule 
& MIN             & MIN $\rightarrow$ CUB & CUB            & CUB $\rightarrow$ MIN   \\
\midrule
Finetuning & \textbf{r=0.82, p=2e-4}  & \textbf{r=0.71, p=3e-3}        & \textbf{r=0.96, p=7e-9} & r=0.28, p=0.31     \\
MAML       & \textbf{r=-0.77, p=8e-4} & \textbf{r=-0.85, p=6e-5}       & r=0.36, p=0.18 & \textbf{r=0.90, p=4e-6}    \\
Reptile    & r=0.27, p=0.3   & r=0.50, p=0.06        & r=0.3, p=0.28  & r=0.31, p=0.27    \\
\bottomrule
\end{tabular}
\end{adjustbox}
\end{table}

The results of this experiment are shown in \autoref{fig:corrs}. 
From this figure, we see that finetuning yields the best joint classification accuracy in all scenarios.
From this figure, we see the following things. 
\begin{itemize}
    \item The within-distribution few-shot learning performance is better than the out-of-distribution performance for all techniques
    \item MAML achieves the best few-shot learning performance when using a shallow backbone (conv-4)
    \item When the backbone becomes deeper, the features learned by MAML become less discriminative
    \item Finetuning learns the most discriminative set of features for direct joint classification on a large set of classes
\end{itemize}
However, we note that the joint classification performance either weakly correlates or does not correlate with the few-shot learning performance across the different techniques.
We note that these correlation patterns may be affected by the fact that we used the best-reported hyperparameters for Reptile for the Conv-4 backbone, while we also use ResNet-10 and ResNet-18 backbones \citep{he2015delving} in different settings. 
For finetuning, however, we do observe an improvement in few-shot learning performance as the backbone becomes deeper.

Next, we investigate whether there are statistically significant relationships per technique between the joint classification accuracy and the few-shot performance.
\autoref{tab:correlations} displays the Pearson correlation and corresponding p-values for individual techniques for the experiment in \autoref{sec:infcont}. 
As we can see, there are strong and significant ($\alpha = 0.005$) correlations between the joint classification accuracy and the few-shot learning performance of finetuning in three settings. 
For MAML, there are strong negative correlations on miniImageNet and miniImageNet $\rightarrow$ CUB, indicating that a lower joint classification accuracy is often associated with better few-shot learning performance.
For Reptile, the correlations are non-significant and mild to weak. 

\section{Conclusion}
\label{sec:discuss}

In this work, we investigated 1)~why MAML and Reptile can outperform finetuning in \emph{within-distribution settings}, and 2)~why finetuning can outperform gradient-based meta-learning techniques such as MAML and Reptile when the test data distribution diverges from the training data distribution.

We have shown how the optimization objectives of the three techniques can be interpreted as maximizing the direct performance, post-adaptation performance, and a combination of the two, respectively. 
That is, finetuning aims to maximize the direct performance whereas MAML aims to maximize the performance \textit{after} a few adaptation steps, making it a look-ahead objective.
Reptile is a combination of the two as it focuses on both the initial performance as well as the performance after every update step on a given task. 
As a result, finetuning will favour an initialization that jointly minimizes the loss function, whereas MAML may settle for an inferior initialization that yields more promising results after a few gradient update steps. 
Reptile picks something in between these two extremes. 
Our synthetic example in \autoref{sec:toy} shows that these interpretations of the learning objectives allow us to understand the chosen initialization parameters. 

Our empirical results show that these different objectives translate into different learned initializations. 
We have shown that MAML and Reptile specialize for adaptation in low-data regimes of the training tasks distribution, which explains why these techniques can outperform finetuning as observed by \citet{chen2019closer,finn2017model,nichol2018reptile}, answering our first research question.
Both the weights of the output layer and the data scarcity in training tasks play an important role in facilitating this specialization, allowing them to gain an advantage over finetuning.

Moreover, we have found that finetuning learns a broad and diverse set of features that allows it to discriminate between many different classes. 
MAML and Reptile, in contrast, optimize a look-ahead objective and settle for a less diverse and broad feature space as long as it facilitates robust adaptation in low-data regimes of the \textit{same} data distribution (as that is used to optimize the look-ahead objective). 
This can explain findings by \citet{chen2019closer}, who show that finetuning can yield superior few-shot learning performance in out-of-distribution settings.
However, we do not observe a general correlation between the feature diversity and the few-shot learning performance across finetuning, Reptile, and MAML.

Another result is that MAML yields the best few-shot learning performance when using the Conv-4 backbone in all settings.  
Interestingly, the features learned by MAML become less discriminative as the depth of the backbone increases.
This may indicate an over-specialization, and it may be interesting to see whether adding a penalty for narrow features may prevent this and increase the few-shot learning performance with deeper backbones and in out-of-distribution settings, which has been observed to be problematic by \citet{rusu2018meta} and \citet{chen2019closer} respectively.  
As this is beyond the scope of our research questions, we leave this for future work. 
Another fruitful direction for future work would be to quantify the distance or similarity between different tasks and to investigate the behaviour of meta-learning algorithms as a function of this quantitative measure. 
An additional benefit of such a measure of task similarity would be that it could allow us to detect when a new task is within-distribution or out-of-distribution, which could inform the choice of which algorithm to use.

In summary, our results suggest that the answer to our second research question is that MAML and Reptile may fail to quickly learn out-of-distribution tasks due to their over-specialization to the training data distribution caused by their look-ahead objective, whereas finetuning learns broad features that allow it to learn new out-of-distribution concepts.
This is supported by the fact that in almost all scenarios, there are statistically significant relationships between the broadness of the learned features and the few-shot learning ability for finetuning.

\section*{Acknowledgements}
This work was performed using the compute resources from the Academic Leiden Interdisciplinary Cluster Environment (ALICE) provided by Leiden University, as well as the Dutch national e-infrastructure with the support of SURF Cooperative.

\section*{Declarations}

\subsection*{Conflicts of Interest}

\paragraph{Funding}
Not applicable: no funding was received for this work.

\paragraph{Employment} All authors declare that there is no recent, present, or anticipated employment by any organization that may gain or lose financially through publication of this manuscript. 

\paragraph{Interests} All authors certify that they have no affiliations with or involvement in any organization or entity with any financial interest or non-financial interest in the subject matter or materials discussed in this manuscript.

\subsection*{Compliance with Ethical Standards}

Not applicable: this research did not involve human participants, nor did it involve animals. 

\subsection*{Consent to participate }

Not applicable. 

\subsection*{Consent for publication}

Not applicable: this research does not involve personal data, and publishing of this manuscript will not result in the disruption of any individual's privacy.

\subsection*{Availability of data and material}

All data that was used in this research have been published as benchmarks by \citet{deng2009imagenet,vinyals2016matching} (miniImageNet) and \citet{wah2011caltech} (CUB), and is publicly available. The data generator for sine wave regression experiments can be found in the provided code (see below).

\subsection*{Code availability}
All code that was used for this research is made publicly available at \url{https://github.com/mikehuisman/revisiting-learned-optimizers}.

\subsection*{Authors' contributions}

MH has conducted the research presented in this manuscript.
AP and JvR have regularly provided feedback on the work, contributed towards the interpretation of results, and have critically revised the whole.    

All authors approve the current version to be published and agree to be accountable for all aspects of the work in ensuring that questions related to the accuracy or integrity of any part of the work are appropriately investigated and resolved.

\subsection*{Ethics approval}
Not applicable. 

\bibliographystyle{abbrvnat}
\bibliography{refs}   

\end{document}